\documentclass[journal,twocolumn,final]{IEEEtran}

\usepackage{amssymb,amsmath}
\usepackage{graphicx}
\usepackage{comment}
\usepackage{cite}
\usepackage{hyperref}
\usepackage{subcaption}
\usepackage[dvipsnames]{xcolor}
\usepackage{bm}
\hypersetup{
  pdftitle={},
  pdfauthor={},
    pdfnewwindow=true,      % links in new window
    colorlinks=true,       % false: boxed links; true: colored links
    linkcolor=blue,          % color of internal links
    citecolor=magenta,        % color of links to bibliography
    filecolor=magenta,      % color of file links
    urlcolor=teal%magenta           % color of external links		
}
\usepackage{tikz}
\usepackage{pgfplots}
\pgfplotsset{compat=newest}
\usetikzlibrary{patterns,shapes,arrows.meta,calc,intersections,backgrounds}
\usetikzlibrary{fillbetween}
\usepgfplotslibrary{fillbetween}

\usepackage{adjustbox}

\pgfmathdeclarefunction{gauss}{3}{%
  \pgfmathparse{1/(#3*sqrt(2*pi))*exp(-((#1-#2)^2)/(2*#3^2))}%
}

\title{On the Geometry of Receiver Operating Characteristic and Precision-Recall Curves}
\author{Reza~Sameni\thanks{ \href{https://orcid.org/0000-0003-4913-6825}{R.~Sameni} is with the Department of Biomedical Informatics, Emory University and the Department of Biomedical Engineering, Georgia Institute of Technology and Emory University. Email: \href{rsameni@dbmi.emory.edu}{rsameni@dbmi.emory.edu}. Web: \url{https://sameni.info/}}\\Version 3.0, April 2026}

\begin{document}

\maketitle

\begin{abstract}
We study the geometry of Receiver Operating Characteristic (ROC) and Precision-Recall (PR) curves in binary classification problems. The key finding is that many of the most commonly used binary classification metrics are merely functions of the composition function $G := F_p \circ F_n^{-1}$, where $F_p(\cdot)$ and $F_n(\cdot)$ are the class-conditional cumulative distribution functions of the classifier scores in the positive and negative classes, respectively. This geometric perspective facilitates the selection of operating points, understanding the effect of decision thresholds, and comparison between classifiers. It also helps explain how the shapes and geometry of ROC/PR curves reflect classifier behavior, providing objective tools for building classifiers optimized for specific applications with context-specific constraints. We further explore the conditions for classifier dominance, present analytical and numerical examples demonstrating the effects of class separability and variance on ROC and PR geometries, and derive a link between the positive-to-negative class leakage function $G(\cdot)$ and the Kullback--Leibler divergence. The framework highlights practical considerations, such as model calibration, cost-sensitive optimization, and operating point selection under real-world capacity constraints, enabling more informed approaches to classifier deployment and decision-making.
\end{abstract}

\section{Introduction}
Classification is a core application of machine learning (ML) and artificial intelligence (AI), with broad use across science, engineering, and medicine. Classifier performance is typically evaluated using metrics derived from \textit{Receiver Operating Characteristic (ROC)} and \textit{Precision-Recall (PR)} curves. While these metrics are widely used, their geometric structure and interrelationships are often overlooked.

A geometric view of these metrics can provide valuable insights, demonstrating how metrics such as sensitivity, specificity, precision, and recall are connected through the distributions of classifier scores and class prevalences.

Motivated by these insights, we study the geometry of ROC and PR curves. We show that many common binary classification metrics naturally arise from a single composition function linking the score distributions across classes, while others are functions of both class-conditional cumulative distribution functions. This yields a unified geometric framework for understanding classifier behavior, comparing models, and selecting optimal operating points for practical applications.

The hereby presented perspective has been partially explored in the literature. Interested readers are invited to refer to these works for further context and insights \cite{krzanowski2009roc,Menon2016BipartiteRanking,Leisman2018,Nahm2022}.

\section{Binary Classification Setup}
Consider a generic binary classification problem in which a classifier $\mathcal{C}$ assigns continuous real-valued scores $x \in \mathbb{R}$ to input sample (data) points. These scores represent the classifier's assessment of how strongly the sample point is associated with either the negative class ($c_n$) or the positive class ($c_p$). The score may or may not have been calibrated to represent actual probabilities of association with the two classes using a given training dataset. The objective is to determine whether a given sample point `$s$' belongs to the positive or negative class based on its score. Here, we focus on \textit{pretrained classifiers}, as our interest is in evaluating classification performance and metrics on validation or test data, without delving into the specifics of the model training process. Nonetheless, we adopt an analytical probabilistic approach that enables the application of our findings to classifier design.

The score $x$ is (generally) a function of the validation/test sample point $s$, i.e., $x(s)$. For notational simplicity, we omit the argument $s$ unless explicitly needed.

\begin{figure}[tb]
\centering
\begin{tikzpicture}
\begin{axis}[
    width=8cm,
    height=5.5cm,
    xlabel={$x$},
    xlabel style={at={(axis description cs:1.01,0)}, anchor=west},
    ylabel={Probability Density},
    legend style={at={(0.97,1.1)}, anchor=north east},
    axis lines=left,
    domain=-5:10,
    samples=400,
    xtick=\empty,
    ytick=\empty,
    clip=false
]
% Baseline for fill (x-axis)
\path[name path=axis] (axis cs:-5,0) -- (axis cs:10,0);
% Define f0
\addplot[name path=f0, orange, thick, domain=-5:10] 
    {0.75 * gauss(x,-1,1.2) + 0.25 * gauss(x,2,0.8)};
\addlegendentry{$f_n(x)$}
% Define f1
\addplot[name path=f1, black, thick, domain=-5:10] 
    {0.35 * exp(-0.5 * (x - 3.5)^2 / 2) * (1 - 0.15*(x - 3.5))};
\addlegendentry{$f_p(x)$}
% Fill under f0: x < \tau (negative region)
\addplot [
    orange!40,
    opacity=0.4
] fill between [
    of=f0 and axis,
    soft clip={domain=-5:1.60}
];
% Fill under f0: x ≥ \tau (false positives)
\addplot [
    orange!70,
    opacity=0.4
] fill between [
    of=f0 and axis,
    soft clip={domain=1.60:10}
];
% Fill under f1: x < \tau (false negatives)
\addplot [
    black!30,
    opacity=0.4
] fill between [
    of=f1 and axis,
    soft clip={domain=-5:1.60}
];
% Fill under f1: x ≥ \tau (true positives)
\addplot [
    black!70,
    opacity=0.4
] fill between [
    of=f1 and axis,
    soft clip={domain=1.60:10}
];
% Threshold line
\addplot[dashed, black] coordinates {(1.60,0) (1.60,0.18)};
\node at (axis cs:1.60,-0.01) [anchor=north] {$\tau$};
% \node at (axis cs:3.2, 0.22) [anchor=north] {$\mathrm{tpr}(\tau)$};
% \node at (axis cs:-1.0, 0.1) [anchor=north] {$\mathrm{tnr}(\tau)$};
\end{axis}
\end{tikzpicture}
\caption{Illustration of score distributions $f_n(x)$ and $f_p(x)$ in a binary classification problem. Shaded regions show classifier decisions based on the threshold $\tau$; observed score distribution $f(x)$ (pool of both classes) not shown. See \eqref{eq:tpr}, \eqref{eq:tnr}, \eqref{eq:fpr}, and \eqref{eq:fnr} for definitions of shaded areas.}
\label{fig:score-distributions}
\end{figure}
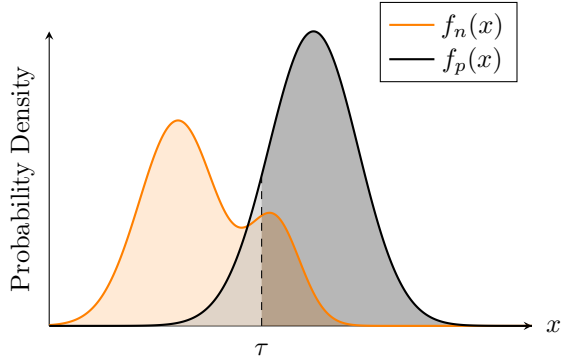

Let $f(x, s)$ denote the joint distribution of scores and sample points for a classifier on a dataset. The score distributions for the evaluated data, conditioned on the negative and positive classes, are denoted by $f_n(x)$ and $f_p(x)$, respectively, and are given by (see Fig.~\ref{fig:score-distributions}):
\begin{equation}
\begin{array}{l}
f_n(x)\!:=\!f(x \mid s \in c_n)\! =\! \displaystyle\frac{f(x, s \in c_n)}{\mathrm{Pr}(s \in c_n)}\!=\!\frac{f(x, s \in c_n)}{\pi_n},\\[1em]
f_p(x)\!:=\!f(x \mid s \in c_p)\!=\! \displaystyle\frac{f(x, s \in c_p)}{\mathrm{Pr}(s \in c_p)}\!=\!\frac{f(x, s \in c_p)}{\pi_p}.
\end{array}
\end{equation}
Here, $\pi_n := \mathrm{Pr}(s \in c_n)$ and $\pi_p := \mathrm{Pr}(s \in c_p)$ represent the prior probabilities of the negative and positive classes, respectively, and satisfy $\pi_n + \pi_p = 1$.

The cumulative distribution functions (CDFs) associated with $f_n(x)$ and $f_p(x)$ are shown as shaded areas in Fig.~\ref{fig:score-distributions}:
\begin{equation}
F_n(x) = \int_{-\infty}^x f_n(t)\, dt, \quad F_p(x) = \int_{-\infty}^x f_p(t)\, dt. 
\end{equation}
Apparently $F_n(\cdot), F_p(\cdot): (-\infty, \infty) \to [0, 1]$.

Our decision rule uses a threshold $\tau$ that partitions the score space into two regions: if $x \geq \tau$, the classifier assigns the sample point to class $c_p$; otherwise, it assigns the sample point to class $c_n$:
\begin{equation}
\delta(s) =
\begin{cases}
c_p, & \text{if } x \geq \tau \\
c_n, & \text{if } x < \tau
\end{cases}
\end{equation}
The choice of threshold $\tau$ directly impacts the trade-offs between true and false positives and negatives, defining the classifier's operating point on the ROC or PR curves.

\section{Binary Classification Metrics}
Given the score distributions $f_n(x)$ and $f_p(x)$ under the negative and positive classes, respectively, the performance of the classifier at a decision threshold $\tau$ can be expressed using standard detection theoretical or ML metrics~\cite{van2004detection}:

\begin{itemize}
    \item True Positive Rate (TPR):
\begin{equation}
\mathrm{tpr}(\tau) := \int_\tau^\infty f_p(x) \, dx = 1 - F_p(\tau)
\label{eq:tpr}
\end{equation}

\item True Negative Rate (TNR):
\begin{equation}
\mathrm{tnr}(\tau) := \int_{-\infty}^\tau f_n(x) \, dx = F_n(\tau)
\label{eq:tnr}
\end{equation}

\item False Positive Rate (FPR):
\begin{equation}
\mathrm{fpr}(\tau) := \int_\tau^\infty f_n(x) \, dx = 1 - F_n(\tau)
\label{eq:fpr}
\end{equation}

\item False Negative Rate (FNR):
\begin{equation}
\mathrm{fnr}(\tau) := \int_{-\infty}^\tau f_p(x) \, dx = F_p(\tau)
\label{eq:fnr}
\end{equation}
\end{itemize}

Equations \eqref{eq:tpr}--\eqref{eq:fnr} quantify how the decision threshold $\tau$ partitions the score space into predicted positive and negative classes. They provide the foundation for evaluating standard binary classification performance metrics, such as precision, recall, specificity, and the areas under the ROC and PR curves.

\subsection{Finite Sample-Size Approximations}
In practice, classifier performance is often evaluated on finite-size datasets. Suppose an evaluation or test set of size $T$, consisting of $P$ positive and $N = T - P$ negative sample points. In this setting, when we apply the classifier $\mathcal{C}$ to the dataset, the true positive, true negative, false positive, and false negative counts at a given threshold $\tau$ are denoted by $\mathrm{TP}(\tau)$, $\mathrm{TN}(\tau)$, $\mathrm{FP}(\tau)$, and $\mathrm{FN}(\tau)$, respectively. These satisfy:
\begin{equation}
\begin{array}{c}
     \mathrm{TP}(\tau) + \mathrm{FN}(\tau) = P,\\
     \mathrm{TN}(\tau) + \mathrm{FP}(\tau) = N,\\
     \mathrm{TP}(\tau) + \mathrm{TN}(\tau) + \mathrm{FP}(\tau) + \mathrm{FN}(\tau) = T.
\end{array}
\label{eq:confusion_matrix}
\end{equation}
Using \eqref{eq:confusion_matrix}, the performance metrics and class prior probabilities can be approximated as:
\begin{equation}
\begin{array}{ll}
     \mathrm{tpr}(\tau) \approx \displaystyle\frac{\mathrm{TP}(\tau)}{P}, & \mathrm{fpr}(\tau) = 1 - \mathrm{tnr}(\tau) \approx \displaystyle\frac{\mathrm{FP}(\tau)}{N},\\[3ex]
     \mathrm{tnr}(\tau) \approx \displaystyle\frac{\mathrm{TN}(\tau)}{N}, & \mathrm{fnr}(\tau) = 1 - \mathrm{tpr}(\tau) \approx \displaystyle\frac{\mathrm{FN}(\tau)}{P},\\[3ex]
    \pi_p \approx \displaystyle\frac{P}{T}, &
    \pi_n = 1 - \pi_p \approx \displaystyle\frac{N}{T}.
    \end{array}
\label{eq:eval_metrics}
\end{equation}
These empirical approximations converge to their distribution-based counterparts as $T \rightarrow \infty$, under the assumption that samples are independently drawn from the underlying score distributions $f_n(x)$ and $f_p(x)$.

\begin{figure}[tb]
    \centering
    \includegraphics[width=\linewidth]{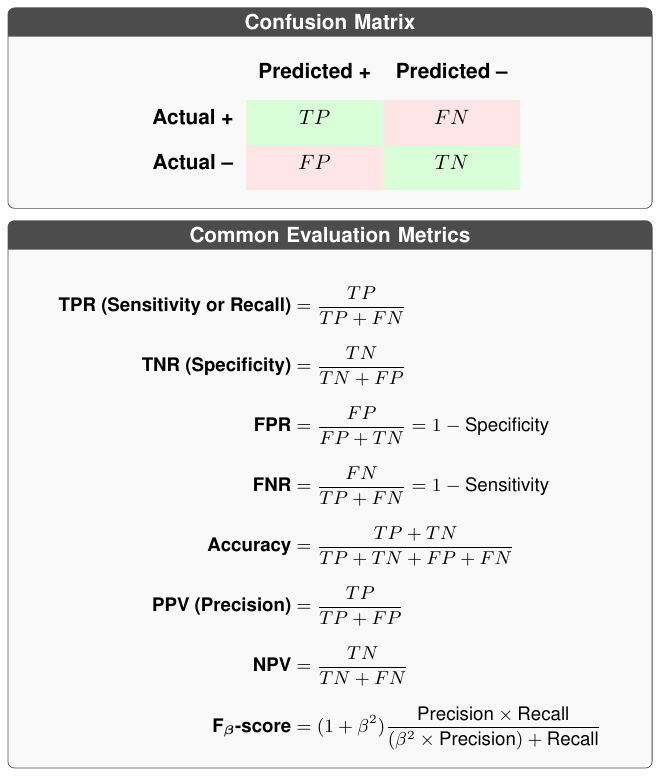}
    \caption{Common classification evaluation measures}
    \label{fig:confusion_matrix}
\end{figure}

The most common classification evaluation measures derived from $\mathrm{TP}$, $\mathrm{TN}$, $\mathrm{FP}$ and $\mathrm{FN}$ are summarized in Fig.~\ref{fig:confusion_matrix}. Note that when a trained classifier is applied to an evaluation or test dataset with fixed values of $P$ and $N$, the resulting confusion matrix is fully determined by the decision threshold. Therefore, despite the diversity of evaluation metrics, a given binary classifier has only a single degree of freedom at inference time---defined by the choice of its decision threshold.
 
\subsection{The Positive-to-Negative Class Leakage Function}
We next introduce a function to which most properties of any binary classifier can be attributed. For any $u \in [0,1]$, $F_n^{-1}(u)$ is a mapping from the negative class probability space back to the score space. In other words, $F_n^{-1}(u)$ provides the score $t$ at which the probability of observing scores given to sample points from class $c_n$ is $u$ (i.e., a fraction $u$ of the negative class distribution lies below $t$). Therefore, applying $F_p$ to this same threshold---i.e., evaluating $F_p\bigl(F_n^{-1}(u)\bigr)$---answers the question: \textit{``Given the threshold where $u$ proportion of scores assigned to negatives lie below, how likely is it that a score assigned to a sample point from the positive class also falls below this same threshold (resulting in its misclassification)?''} A good classifier would make this probability negligible, so that a threshold accommodating a certain portion of negatives does not include too many positives. With this insight, we define
\begin{equation}
\boxed{
 G(u) := F_p\bigl(F_n^{-1}(u)\bigr) = (F_p \circ F_n^{-1})(u),
}
\label{eq:G}
\end{equation}
as the \textit{positive-to-negative class leakage function}, since it measures how much the negative threshold at level $u$ encroaches upon the positive distribution. $G(\cdot)$ is a monotonic mapping between two probability spaces: $G(\cdot): [0, 1] \to [0, 1]$ and it fulfills $G(0) = 0$ and $G(1) = 1$ with a probability density function:
\begin{equation}
    g(u) := \frac{d}{du}G(u)
\label{eq:G_density}
\end{equation}

Intuitively, the more $f_n(\cdot)$ and $f_p(\cdot)$ overlap, the larger the area under $G(\cdot)$ becomes, indicating that many positive sample points are scored below a threshold chosen for negatives. We will provide supporting evidence for this property in Section~\ref{sec:classifier_metrics}.

% ANOTHER POTENTIAL INTERPRETATIONS: A well-known result from probability theory, known as the \textit{probability integral transform}, states that if a continuous random variable $x$ has CDF $F_x(x)$, then the transformed variable $u = F_x(x)$ is uniformly distributed over the interval $[0, 1]$, i.e., $u \sim \mathrm{Uniform}(0, 1)$~\cite{papoulis2002probability}. Conversely, if $u \sim \mathrm{Uniform}(0, 1)$, then applying the inverse CDF yields $x = F_x^{-1}(u)$, which produces a new random variable $x$ that follows the original distribution associated with $F_x$. Applying this property to the composition function $G(x) = (F_p \circ F_n^{-1})(x)$ in \eqref{eq:auroc} provides another interpretation: 
% \textcolor{blue}{Needs more work, but if $x$ is uniform, $F_n^{-1}(x)$ takes the pdf of $c_n$; then we apply $F_p$ on it; then what?!...}

\section{Classifier Evaluation Measures}
\label{sec:classifier_metrics}
\subsection{Receiver Operating Characteristic (ROC) Curve}
The ROC curve is a graphical representation showing TPR (i.e., $\mathrm{sensitivity}$, or $1-\mathrm{miss\!\!-\!\!rate}$) against FPR (i.e., $1-\mathrm{specificity}$), illustrating the trade-off between false positives and true positives as the decision threshold~$\tau$ varies~\cite{krzanowski2009roc}.

According to \eqref{eq:tpr} and \eqref{eq:fpr}, TPR and FPR are monotonic functions of $\tau$. Therefore, each threshold $\tau$ corresponds to unique TPR and FPR values:
\begin{equation}
    \tau = F_p^{-1} \bigl(1 - \mathrm{tpr}(\tau)\bigr) = F_n^{-1} \bigl(1 - \mathrm{fpr}(\tau)\bigr),
\end{equation}
which results in $\mathrm{tpr}(\tau) = 1 - (F_p \circ F_n^{-1}) \bigl(1 - \mathrm{fpr}(\tau)\bigr)$ providing a compact formulation of the ROC curve~\cite[Sec~2.2.4]{krzanowski2009roc}:
\begin{equation}
\boxed{
    \mathrm{tpr}(\tau) = 1 - G\bigl(1 - \mathrm{fpr}(\tau)\bigr)
    }
\label{eq:roc_curve}
\end{equation}
parameterized by the decision threshold $\tau$. From \eqref{eq:roc_curve}, the monotonically increasing shape of the ROC curve is apparent; it starts at $(0,0)$ and ends at $(1,1)$. It also leads to the following results:
\begin{equation}
    \mathrm{miss\!\!-\!\!rate}(\tau) = 1 - \mathrm{tpr}(\tau) = G \bigl(\mathrm{specificity}(\tau)\bigr),
\label{eq:specificity_missrate}
\end{equation}

\begin{equation}
\mathrm{fpr}(\tau) = 1 - G^{-1}\bigl(1 - \mathrm{tpr}(\tau)\bigr).
\label{eq:roc_inverse}
\end{equation}

An important implication of \eqref{eq:roc_curve} is that as long as $F_p$ and $F_n$ are well-learned from the training data (or known/presumed a priori), the shape of the ROC curve is not impacted by the positive vs. negative class prevalence in the validation/test sets. We will later show that this is not the case for the PR curve.

\subsection{Area Under the ROC Curve (AUROC)}
\label{sec:auroc}
From \eqref{eq:roc_curve}, the area under the ROC curve (AUROC) is:
\begin{equation}
\mathrm{AUROC} = \!\!\int_{0}^{1} \mathrm{tpr}(u) \, du = \!\!\int_{0}^{1} \left[1 - (F_p \circ F_n^{-1})(1 - u) \right] du,
\end{equation}
With a change of variable $v = 1 - u$ and $du = -dv$, we find:
\begin{equation}
     \mathrm{AUROC} = \displaystyle\int_{1}^{0} \left[1 - G(v) \right] (-dv) = 1 - \displaystyle\int_{0}^{1} G(v) dv.
\label{eq:auroc}
\end{equation}
This shown how the AUROC is fully determined by the composition of the class-conditional score distribution CDFs. Accordingly, smaller areas under $G(\cdot)$ indicate less overlap between the positive and negative class score ranges, resulting in higher AUROC values (indicating stronger class separation). In other words, maximizing AUROC---which is the objective function used in many classifier design strategies---minimizes the area under $G(\cdot)$.

\paragraph{Alternative Interpretation} Equation \eqref{eq:auroc} aligns with the interpretation of AUROC as the probability of a randomly chosen positive sample point ranking higher than a randomly chosen negative one. As shown in the Appendix~\cite{Hanley1982,Fawcett2006}: 
\begin{equation}
\begin{array}{rl}
\mathrm{AUROC} &= \Pr(x_p > x_n)\\[0.5em] &= \displaystyle\int_{-\infty}^{\infty} f_p(x)\, F_n(x)\, dx = \displaystyle\int_{-\infty}^{\infty} \dot{F}_p(x)\, F_n(x)\, dx
\end{array}
\label{eq:auroc_prob}
\end{equation}
where \(x_p \sim f_p(x)\) is a score randomly drawn from the positive class score distribution, and \(x_n \sim f_n(x)\) is a score randomly drawn from the negative class score distribution. The two are assumed to be independent. Combining \eqref{eq:auroc} and \eqref{eq:auroc_prob} yields
\begin{equation}
\Pr(x_n \geq x_p) = \displaystyle\int_{0}^{1} G(v)\, dv
\label{eq:prob_G}
\end{equation}
which is the probability that an ideal classifier seeks to minimize.

A typical ROC curve with the highlighted AUROC and the composition function $G(\cdot)$ is shown in Fig.~\ref{fig:typical_roc_curve}. In Appendix~\ref{sec:kl_leakage_derivation}, we further relate $G(\cdot)$ with the Kullback–Leibler divergence between the class score density functions $f_p$ and $f_n$.

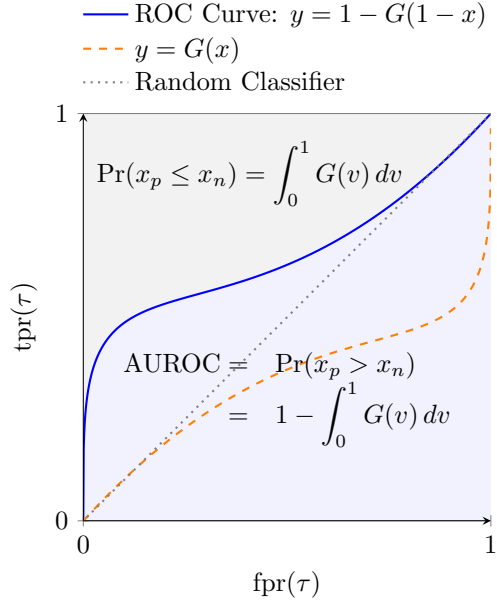
\begin{figure}[tb]
\centering
\begin{tikzpicture}
\begin{axis}[
    width=7cm,
    height=7cm,
    xlabel={$\mathrm{fpr}(\tau)$},
    ylabel={$\mathrm{tpr}(\tau)$},
    axis lines=left,
    xmin=0, xmax=1,
    ymin=0, ymax=1,
    xtick={0,1},
    ytick={0,1},
    axis on top,
    legend style={
        at={(0.5,1.3)},
        anchor=north,
        draw=none,
        legend cell align=left
    }
]

% ROC Curve
\addplot[
    name path=roc,
    thick,
    blue,
    domain=0:1,
    samples=1000
] {(x)^(0.25) + 0.85*((0.5 - x)^2 - (0.5)^2)};
\addlegendentry{ROC Curve: $y = 1 - G(1 - x)$}

% Top boundary (y=1)
\addplot[
    name path=top,
    draw=gray,
    forget plot
] coordinates {(0,1) (1,1)};

% Bottom boundary (y=0)
\addplot[
    name path=bottom,
    draw=gray,
    forget plot
] coordinates {(0,0) (1,0)};

% Shaded area above ROC
\addplot[
    gray!10,
    forget plot
] fill between[of=roc and top];

% Shaded area under ROC
\addplot[
    blue!5,
    forget plot
] fill between[of=roc and bottom];

% Additional plot y=G(x)
\addplot[
    thick,
    dashed,
    orange,
    domain=0:1,
    samples=1000,
    legend image post style={thick, dashed, orange}
] {1 - (1 - x)^(0.25) - 0.85*((x - 0.5)^2 - (0.5)^2)};
\addlegendentry{$y = G(x)$}

% Random Classifier line
\addplot[
    thick,
    dotted,
    gray,
    legend image post style={thick, dotted, gray}
] {x};
\addlegendentry{Random Classifier}

% Integral labels
\node at (axis cs:0.01,0.75) [anchor=south west] {$\Pr(x_p \leq x_n) = \displaystyle\int_0^1 G(v)\,dv$};

\node at (axis cs:0.50,0.3) {$\begin{array}{rl}
     \mathrm{AUROC} = & \Pr(x_p > x_n)\\
      = & 1 - \displaystyle\int_0^1 G(v)\,dv
\end{array}$};
% \node at (axis cs:0.3,0.1) {$y = G(x)$};

\end{axis}
\end{tikzpicture}
\caption{A typical ROC curve highlighting the composition function $G(\cdot)$ and the AUROC.}
\label{fig:typical_roc_curve}
\end{figure}

\subsection{Precision-Recall (PR) Curve}
The PR curve shows precision (or the positive predictive value: PPV),  vs.\ recall (sensitivity or TPR) ~\cite{Davis2006}, illustrating the potential trade-off between the accuracy of positive predictions and the ability to detect positive cases as the decision threshold~$\tau$ is changed.

Probabilistically, PPV at threshold~$\tau$ is $\Pr(s \in c_p \mid x > \tau)$, which using the Bayes' rule can be expressed in terms of the TPR, score distributions and class priors:
\begin{equation}
\begin{array}{rl}
    \mathrm{ppv}(\tau) & = \displaystyle\frac{\Pr(x > \tau \mid s \in c_p) \cdot \Pr(s \in c_p)}
         {\Pr(x > \tau)} \\[3ex]    
    & = \displaystyle\frac{\pi_p \displaystyle\int_\tau^\infty f_p(x) \, dx}
           {\pi_p \displaystyle\int_\tau^\infty f_p(x) \, dx + \pi_n \displaystyle\int_\tau^\infty f_n(x) \, dx}\\[5ex]
    & = \displaystyle\frac{\pi_p \cdot [1 - F_p(\tau)]}
           {\pi_p \cdot [1 - F_p(\tau)] + \pi_n \cdot [1 - F_n(\tau)]} \\[3ex]
% \end{array}
% \label{eq:ppv_integral_form}
% \end{equation}
% PPV can be expressed in terms of the score distributions and class priors as:
% \begin{equation}
% \begin{array}{rl} \mathrm{ppv}(\tau)
    & = \displaystyle\frac{\pi_p \cdot \mathrm{tpr}(\tau)}{\pi_p \cdot \mathrm{tpr}(\tau) + \pi_n \cdot \mathrm{fpr}(\tau)} = \displaystyle\frac{1}{1 + \displaystyle\frac{\pi_n}{\pi_p} \cdot \frac{\mathrm{fpr}(\tau)}{\mathrm{tpr}(\tau)}}\\[3ex]% = \displaystyle \frac{1}{1 + \frac{\pi_n}{\pi_p} \cdot \left[\frac{1 - F_n(\tau)}{1 - F_p(\tau)}\right]}\\
   & = \displaystyle\frac{\mathrm{tpr}(\tau)}{\mathrm{tpr}(\tau) + \displaystyle\frac{\pi_n}{\pi_p} \cdot [1 - (F_n \circ F_p^{-1}) \bigl(1 - \mathrm{tpr}(\tau)\bigr)] }
\end{array}
\label{eq:prc_curve_derivation}
\end{equation}
Therefore
\begin{equation}
\boxed{
\mathrm{ppv}(\tau) = \displaystyle\frac{\mathrm{tpr}(\tau)}{\mathrm{tpr}(\tau) + \displaystyle\frac{\pi_n}{\pi_p} \cdot [1 - G^{-1}\bigl(1 - \mathrm{tpr}(\tau)\bigr)] }
    }
\label{eq:prc_curve}    
\end{equation}
Accordingly, the precision and recall curve depends not only on the classifier's score distributions but also on the class prior probabilities in the dataset. As compared with the ROC curve, it is also a more complex function of $G(\cdot)$.

\subsubsection{Start and End Points of the PR Curve}
From \eqref{eq:prc_curve}, as $\tau \to -\infty$ we have $\mathrm{tpr}(\tau)=1$, and the PR curve always ends at $(1, \pi_p)$. However, the starting point of the PR curve is not as evident because as $\tau \to +\infty$, $\mathrm{tpr}(\tau)\to 0$ and $\mathrm{ppv}(\tau)$ becomes indeterminate ($0/0$). Applying L'Hôpital's rule and using the derivative of $G(\cdot)$ from \eqref{eq:G_derivative}, we can show that the PR curve starts at:
\begin{equation}
    (0, \frac{\mathrm{LR}(\infty)}{\mathrm{LR}(\infty) + 1})
\label{eq:pr_start}
\end{equation}
where $\mathrm{LR}(\infty) 
\;=\; 
\lim_{\tau \to +\infty} \frac{\pi_p f_p(\tau)}{\pi_n f_n(\tau)}$ is the prevalence-adjusted tail likelihood ratio, i.e., the ratio of the positive over negative class contributions to the total probability ($\pi_p\, f_p(\tau) + \pi_n\, f_n(\tau)$) at the right tail of the distribution in Fig.~\ref{fig:score-distributions}. Equation \eqref{eq:pr_start} can also be expanded into a more intuitive form\footnote{Equation \eqref{eq:pr_start_simplified} can alternatively be obtained by applying L'Hôpital's rule directly to the second line of \eqref{eq:prc_curve_derivation} as $\tau \to +\infty$.}:
\begin{equation}
    \left(0,\; \lim_{\tau \to +\infty} 
    \frac{\pi_p\, f_p(\tau)}{\pi_p\, f_p(\tau) + \pi_n\, f_n(\tau)}\right)
\label{eq:pr_start_simplified}
\end{equation}
Using the Bayes' theorem, the fraction in \eqref{eq:pr_start_simplified} is, in fact, the posterior probability of the sample point being a true positive given a score of $\tau$, i.e., $\Pr(s\in c_p | \mathrm{score} = \tau)$. Therefore, in the limit ($\tau\to+\infty$), which corresponds to the starting point of the PR curve, we are asking: \textit{``As we push the decision threshold to its extreme, what is the probability that a sample point flagged as positive is truly positive?''} 

Therefore, the starting point of the PR curve, as we will show in the case study examples in Section~\ref{sec:case_studies}, can be anywhere between 0 and 1 depending on the class prevalences and tail behavior of $f_p(\tau)$ and $f_n(\tau)$.

\subsection{Accuracy}
The accuracy of a binary classifier at decision threshold $\tau$ is the probability of correct classification, computed as the weighted sum of the true positive and true negative rates (cf. Fig.~\ref{fig:confusion_matrix} for the finite-sample approximation):
\begin{equation}
\begin{array}{rl}
\mathrm{Accuracy}(\tau) &= \pi_p \cdot \mathrm{tpr}(\tau) + \pi_n \cdot \mathrm{tnr}(\tau)\\[0.2em]
&= \pi_p \cdot [1 - F_p(\tau)] + \pi_n \cdot F_n(\tau)
\end{array}
\end{equation}
This shows that accuracy depends on the class priors $\pi_n$ and $\pi_p$, as well as the CDFs of the classifier scores under the positive and negative classes. Apparently, accuracy is not only a function of the composition $G(\cdot)$ but rather a function of both $F_n$ and $F_p$. 

\subsection{\texorpdfstring{\boldmath{$F_\beta$}}{}  Score --- A Generalized Harmonic Metric}
The $F_\beta$ score is a generalization of the $F_1$ score that provides a tunable trade-off between precision and recall. It is the weighted harmonic mean of precision (positive predictive value) and recall (TPR or sensitivity, cf. Fig.~\ref{fig:confusion_matrix}):
\begin{equation}
F_\beta = (1 + \beta^2) \cdot \frac{\mathrm{Precision} \cdot \mathrm{Recall}}{(\beta^2 \cdot \mathrm{Precision}) + \mathrm{Recall}}
\end{equation}
The parameter $\beta > 0$ controls the balance:
\begin{itemize}
    \item $\beta = 1$ yields the standard $F_1$ score, equally weighting precision and recall.
    \item $\beta > 1$ emphasizes recall (sensitivity) more heavily than precision.
    \item $\beta < 1$ emphasizes precision over recall.
\end{itemize}
To relate the $F_\beta$ score to the score distributions and the class priors, we use:

\begin{equation}
\begin{array}{rl}
\mathrm{Precision}(\tau) &= \displaystyle\frac{\pi_p \cdot \mathrm{tpr}(\tau)}{\pi_p \cdot \mathrm{tpr}(\tau) + \pi_n \cdot \mathrm{fpr}(\tau)}\\[1em] \mathrm{Recall}(\tau) &= \mathrm{tpr}(\tau)
\end{array}
\end{equation}
Substituting into the $F_\beta$ formula yields:
\begin{equation}
F_\beta(\tau) = \frac{(1 + \beta^2) \cdot \mathrm{tpr}(\tau)}{
\beta^2 + \mathrm{tpr}(\tau) + \dfrac{\pi_n}{\pi_p} \mathrm{fpr}(\tau)},
\label{eq:f_beta_tpr_fpr}
\end{equation}
which, using the definitions of $\mathrm{tpr}(\tau)$ and $\mathrm{fpr}(\tau)$, \eqref{eq:f_beta_tpr_fpr} can be rewritten as:
\begin{equation}
F_\beta(\tau) = \frac{(1 + \beta^2)[1 - F_p(\tau)]}{ \beta^2 +
[1 - F_p(\tau)] +  \dfrac{\pi_n}{\pi_p} [1 - F_n(\tau)]}
\label{eq:f_beta_F0_F1}
\end{equation}
This shows how $F_\beta$ depends on the threshold $\tau$, the class priors $\pi_n / \pi_p$, and the score distributions $F_n(\tau)$ and $F_p(\tau)$. Larger values of $1 - F_p(\tau)$ (true positives) improve $F_\beta$, while higher $1 - F_n(\tau)$ (false positives) and stronger class imbalance degrade it---especially when $\beta$ emphasizes recall. Similar to accuracy, $F_\beta(\tau)$ is a function of both $F_n(\cdot)$ and $F_p(\cdot)$, not just the composite $G(\cdot)$

\section{Dominating Classifiers}
Suppose we have classifiers $\mathcal{C}_1$ and $\mathcal{C}_2$ with positive-to-negative class leakage functions $G_1(\cdot)$ and $G_2(\cdot)$, respectively. We say that $\mathcal{C}_1$ dominates $\mathcal{C}_2$ if the ROC and PR curves of $\mathcal{C}_1$ lie entirely on or above those of $\mathcal{C}_2$. That is, $\mathcal{C}_1$ achieves sensitivity (TPR) greater than that of $\mathcal{C}_2$ at all false positive rates (FPR), and precision greater than that of $\mathcal{C}_2$ at all recall levels. We refer to this case as \textit{global dominance}. In contrast, if $\mathcal{C}_1$ only outperforms $\mathcal{C}_2$ over a subset of the domain---e.g., at certain FPR ranges or specific recall intervals---we say that $\mathcal{C}_1$ has \textit{local dominance} over $\mathcal{C}_2$. From~\eqref{eq:roc_curve} and~\eqref{eq:prc_curve}, it follows that for both ROC and PR curves local dominance holds whenever, for a given $u$,
\begin{equation}
    G_1(u) < G_2(u)
\end{equation}
and global dominance holds when the above holds for all $u \in [0,1]$. We further name the dominance \textit{non-strict} if $G_1(u) \leq G_2(u)$. Importantly,
\begin{itemize}
    \item When the ROC or PR curve of $\mathcal{C}_1$ globally dominates that of $\mathcal{C}_2$, the corresponding PR or ROC curve is also guaranteed to be globally dominated (since the dominance holds for all $u \in [0,1]$). However, if the dominance is only local, there is no such guarantee---$\mathcal{C}_1$ may outperform $\mathcal{C}_2$ in specific segments of the curves.
    \item Even if a classifier globally or locally dominates another in terms of both ROC and PR curves, there is no guarantee that it also dominates in measures such as accuracy and $F_\beta$, which depend on both $F_p(\cdot)$ and $F_n(\cdot)$, not merely on $G_1(\cdot)$ and $G_2(\cdot)$. In fact, two classifiers can have identical ROC/PR curves but different accuracies or $F_\beta$ scores.
    \item Classification evaluation metrics are statistical aggregations over a population of sample points and do not provide much information about specific sample points or subpopulations. Therefore, in heterogeneous datasets, a classifier that generally underperforms across the full population might still perform well on specific subpopulations, and a classifier that strictly dominates other classifiers may still underperform on specific subgroups.
    \item Even a weak classifier that performs well only on specific subgroups can be beneficial in ensemble methods such as voting or boosting, which combine the strengths of multiple classifiers to improve overall performance.
\end{itemize}

\section{Design Constraints and Operating Point Selection}
Classification is inherently probabilistic, and no practical classifier is ever perfect (unless for trivial classification problems). Therefore, false positives and negative cases are inevitable and depending on the chosen score threshold, different trade-offs arise. Although classifiers are often designed and compared using abstract metrics such as AUROC, accuracy, or $F_{\beta}$ score, real-world constraints can dictate the operating point---that is, the threshold at which a model labels an instance positively or negatively. 

\begin{figure}[tb]
\centering
\begin{tikzpicture}
\begin{axis}[
    width=7cm,
    height=7cm,
    xlabel={$\mathrm{fpr}(\tau)$},
    ylabel={$\mathrm{tpr}(\tau)$},
    axis lines=left,
    xmin=0, xmax=1,
    ymin=0, ymax=1,
    xtick={0,1},
    ytick={0,1},
    axis on top,
    legend style={
        at={(0.78,1.15)},
        anchor=north,
        draw=none,
        legend cell align=left
    }
]

% ROC Curve
\addplot[
    name path=roc,
    thick,
    blue,
    domain=0:1,
    samples=1000
] {(x)^(0.25)};
% \addlegendentry{ROC Curve}

% Top boundary (y=1)
\addplot[
    name path=top,
    draw=gray,
    forget plot
] coordinates {(0,1) (1,1)};

% Bottom boundary (y=0)
\addplot[
    name path=bottom,
    draw=gray,
    forget plot
] coordinates {(0,0) (1,0)};

% Capped positives
\addplot[
    name path=cap_pos,
    brown
] coordinates {(0,0.8) (0.5,0)};

% Max missing cost
\addplot[
    name path=cap_cost,
    purple
] coordinates {(0,0.3) (1,1)};

% Shaded area above ROC
\addplot[
    gray!40,
    opacity=0.4,
    forget plot
] fill between[of=cap_pos and bottom];

% Shaded area under ROC
\addplot[
    blue!20,
    opacity=0.4,
    forget plot
] fill between[of=cap_cost and top];

% Random Classifier line
\addplot[
    thick,
    dotted,
    gray,
    legend image post style={thick, dotted, gray}
] {x};
% \addlegendentry{Random Classifier}

% Fixed specificity (vertical line at x = 0.3)
\addplot[
    thick,
    dashed,
    red
] coordinates {(0.3,0) (0.3,1)};

% Fixed sensitivity (horizontal line at y = 0.7)
\addplot[
    thick,
    dashed,
    OliveGreen
] coordinates {(0,0.7) (1,0.7)};

\node[blue, rotate=20.5] at (axis cs:0.6,0.91) {\scriptsize{ROC Curve}};

\node[brown, rotate=-58] at (axis cs:0.42,0.18) {\scriptsize{Capped Positives}};

\node[gray, rotate=45] at (axis cs:0.55,0.5) {\scriptsize{Random Classifier}};

\node[purple, rotate=35.0] at (axis cs:0.27,0.45) {\scriptsize{Capped FPR \& FNR Cost}};

\node[red, rotate=90] at (axis cs:0.33,0.75) {\scriptsize{Fixed Specificity}};

\node[OliveGreen] at (axis cs:0.78,0.73) {\scriptsize{Fixed Sensitivity}};

\end{axis}
\end{tikzpicture}
\caption{Illustration of different constrained scenarios and the corresponding design regions in the ROC plane.}
\label{fig:roc_design_plane}
\end{figure}
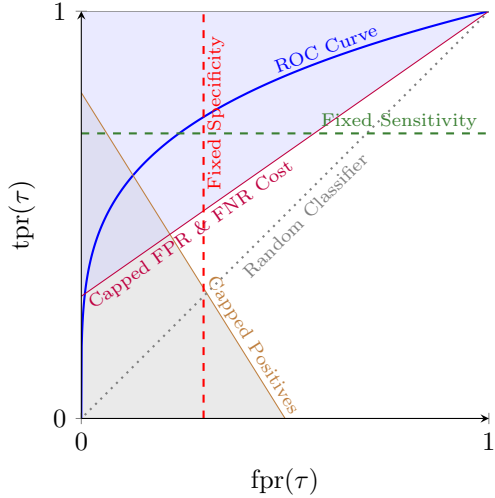

In this section, we explore practical scenarios, illustrating how real-world constraints guide the selection of the operating point. The studied cases and the ROC plane design regions corresponding to each case are illustrated in Fig.~\ref{fig:roc_design_plane}.

\subsection{Relationship between AUROC, Sensitivity and Specificity}
A common practical scenario is when target sensitivity and specificity are specified for a classification problem, and one seeks to design a classifier to meet those conditions. Denoting sensitivity at a target score threshold $\tau$ by $\mathrm{se}$ (or $\mathrm{tpr}(\tau)$) and specificity by $\mathrm{sp}$ (or $1 - \mathrm{fpr}(\tau)$), the corresponding operating point on the ROC curve is $(1-\mathrm{sp}, \mathrm{se})$.

Without additional assumptions on the ROC curve (beyond monotonicity), AUROC is not uniquely determined by the operating point. However, it is tightly bounded. As shown in Fig.~\ref{fig:roc_auc_bounds}, the minimum and maximum achievable AUROC consistent with this operating point are
\begin{equation}
\mathrm{AUROC}_{\min} = \mathrm{se}\cdot\mathrm{sp},
\label{eq:auc_min}
\end{equation}
\begin{equation}
\mathrm{AUROC}_{\max} = \mathrm{se} + \mathrm{sp} - \mathrm{se}\cdot\mathrm{sp}.
\label{eq:auc_max}
\end{equation}
Therefore,
\begin{equation}
\mathrm{se}\,\mathrm{sp} \;\le\; \mathrm{AUROC} \;\le\; \mathrm{se} + \mathrm{sp} - \mathrm{se}\cdot\mathrm{sp}.
\label{eq:auc_bounds}
\end{equation}

The lower bound corresponds to a ROC curve that stays close to the axes for as long as possible, while the upper bound corresponds to a curve that rises as early as possible. The gap between these bounds is
\begin{equation}
\begin{array}{rl}
\mathrm{AUROC}_{\text{range}} &:= \mathrm{AUROC}_{\max} - \mathrm{AUROC}_{\min}\\
&\;= \mathrm{se}(1-\mathrm{sp}) + \mathrm{sp}(1-\mathrm{se}),
\end{array}
\label{eq:auc_range}
\end{equation}
which quantifies the uncertainty in AUROC given a target operating point. Accordingly, the closer the operating point is to the ideal sensitivity and specificity of one, the narrower the range of AUROC becomes. Therefore, designing a classifier that improves both sensitivity and specificity, enforces tighter bounds on the ROC curve.

A useful intermediate case is the triangular ROC approximation, where the curve is approximated by two line segments connecting $(0,0)$ to $(1-\mathrm{sp}, \mathrm{se})$ and then to $(1,1)$. In this case, AUROC is
\begin{equation}
\overline{\mathrm{AUROC}} = \frac{\mathrm{se} + \mathrm{sp}}{2}.
\end{equation}
This value lies exactly at the midpoint of the feasible interval:
\begin{equation}
\overline{\mathrm{AUROC}}
= \frac{\mathrm{AUROC}_{\min} + \mathrm{AUROC}_{\max}}{2}.
\end{equation}

For example, for a sensitivity of $0.8$ and specificity of $0.75$, the operating point on the ROC curve is $(0.25, 0.8)$. This does not uniquely determine the AUROC, but using \eqref{eq:auc_min} and \eqref{eq:auc_max} it constrains it between a minimum and maximum of $0.60$ and $0.95$, with an average AUROC of $0.775$.

Overall, while a target sensitivity and specificity do not determine AUROC, they constrain it to a bounded interval. See \cite{Nahm2022} for further insights on the geometric relationship between the ROC curve and the operating point.

\begin{figure}[tb]
\centering
\begin{tikzpicture}
\begin{axis}[
    width=8cm,
    height=8cm,
    xlabel={$ \mathrm{fpr}(\tau) $},
    ylabel={$ \mathrm{tpr}(\tau) $},
    axis lines=left,
    xmin=0, xmax=1,
    ymin=0, ymax=1,
    xtick={0,0.2,1},
    xticklabels={$0$,$1-\mathrm{sp}$,$1$},
    ytick={0,0.8,1},
    yticklabels={$0$,$\mathrm{se}$,$1$},
    axis on top,
    legend style={
        at={(0.5,-0.18)},
        anchor=north,
            draw=none,
        legend columns=1,
        legend cell align=left
    }
]

% Parameters for the illustrative operating point:
% (1-sp, se) = (0.2, 0.8)

% Minimum-case ROC
\addplot[
    name path=rocmin,
    very thick,
    Maroon
] coordinates {(0,0) (0.2,0) (0.2,0.8) (1,0.8) (1,1)};
\addlegendentry{Minimum AUROC}

% Average / triangular ROC
\addplot[
    name path=roctri,
    very thick,
    NavyBlue
] coordinates {(0,0) (0.2,0.8) (1,1)};
\addlegendentry{Triangular AUROC}

% Maximum-case ROC
\addplot[
    name path=rocmax,
    very thick,
    OliveGreen
] coordinates {(0,0) (0,0.8) (0.2,0.8) (0.2,1) (1,1)};
\addlegendentry{Maximum AUROC}

% Bottom boundary
\addplot[
    name path=bottom,
    draw=none,
    forget plot
] coordinates {(0,0) (1,0)};

% Shade AUROC_min
\addplot[
    Maroon!35,
    opacity=0.45,
    forget plot
] fill between[of=rocmin and bottom];

% Shade incremental area from min to average
\addplot[
    NavyBlue!25,
    opacity=0.45,
    forget plot
] fill between[of=roctri and rocmin];

% Shade incremental area from average to max
\addplot[
    OliveGreen!25,
    opacity=0.45,
    forget plot
] fill between[of=rocmax and roctri];

% Random classifier
\addplot[
    thick,
    dotted,
    gray
] {x};
\addlegendentry{Random classifier}

% Vertical and horizontal guides through the operating point
\addplot[
    thick,
    dashed,
    red
] coordinates {(0.2,0) (0.2,0.8)};
\addlegendentry{$\mathrm{fpr}(\tau)=1-\mathrm{sp}$}

\addplot[
    thick,
    dashed,
    orange
] coordinates {(0,0.8) (0.2,0.8)};
\addlegendentry{$\mathrm{tpr}(\tau)=\mathrm{se}$}

% Operating point
\addplot[
    only marks,
    mark=*,
    mark size=2.2pt,
    black
] coordinates {(0.2,0.8)};
\node[anchor=north west] at (axis cs:0.2,0.8) {\scriptsize $(1-\mathrm{sp}\cdot\mathrm{se})$};

% Labels inside the panel
\node[Maroon] at (axis cs:0.63,0.18) {\scriptsize $\mathrm{AUROC}_{\min}=\mathrm{se}\cdot\mathrm{sp}$};

\node[NavyBlue] at (axis cs:0.76,0.85) {\scriptsize $\overline{\mathrm{AUROC}}=\dfrac{\mathrm{se}+\mathrm{sp}}{2}$};

\node[OliveGreen] at (axis cs:0.3,1.93) {\scriptsize $\mathrm{AUROC}_{\max}=\mathrm{se}+\mathrm{sp}-\mathrm{se}\,\mathrm{sp}$};

\node[gray, rotate=45] at (axis cs:0.35,0.3) {\scriptsize Random classifier};

\end{axis}
\end{tikzpicture}
\caption{Illustration of the minimum, triangular, and maximum ROC curves consistent with a fixed operating point $(1-\mathrm{sp}, \mathrm{se})$. The shaded regions show the corresponding AUROC values. In this example, the operating point is shown at $(0.2,0.8)$, corresponding to sensitivity of $\mathrm{se}=0.8$ and specificity of $\mathrm{sp}=0.8$.}
\label{fig:roc_auc_bounds}
\end{figure}

\subsection{Limited/Fixed Capacity of Admitting Positive Cases}
In real-world scenarios, AI- or machine-learning-powered tools are often integrated into existing human-centric workflows. For example, if a classifier is used for prescreening patients for a specific medical condition, the positively classified cases are typically referred to a physician or a lab test. However, healthcare facilities usually do not have the capacity to accommodate all referrals. They are constrained by the number of available physicians---whether on duty or on call---as well as hospital beds, lab testing capacity and other factors. On the other hand, under-referral (i.e., too few positive referrals) is another undesired scenario that can result in missed diagnoses, delayed treatments, and potential harm to patients, ultimately compromising clinical outcomes and leading to unjustifiable operational costs for a healthcare system.

As a result, there is often a hard or soft limit on the maximum (or even minimum) number of positive cases beyond which the healthcare system cannot admit additional patients. For example, for the maximum, the sum of the true and false positives may be capped, i.e., $\mathrm{TP} + \mathrm{FP} \leq M$, where $M$ is \textit{``the maximum number of cases that the classifier can label as positive while remaining within the healthcare system's capacity for patient admission ($M\leq$P).''} Importantly, these capacities and limitations are dynamic and vary over time and setting, making the problem more complicated. Similar constraints arise in other application domains as well.

Note that in the scenario described above, $M$ may or may not scale with the total number of subjects ($T$). For example, in an overwhelmed healthcare setting operating at maximum capacity, $M$ may not scale easily with an increase in the number of patients ($T$). Defining $m := M/T$, as the \textit{maximum admission rate} and using \eqref{eq:eval_metrics}, we obtain:
\begin{equation}
    \pi_p \mathrm{tpr}(\tau) + \pi_n \mathrm{fpr}(\tau) \leq m
\label{eq:max_referral}
\end{equation}
In the ROC plane, \eqref{eq:max_referral} describes a triangular region below a line with negative slope $-\pi_n/\pi_p=-N/P$, intersecting the TPR (vertical) axis at $m/\pi_p = M/P$ and the FPR (horizontal) axis at $m/\pi_n = M/N$, as shown in Fig.~\ref{fig:roc_design_plane}. Apparently feasible operating points of the classifier must lie within or below this region to satisfy the capacity constraint. In this scenario, the ``optimal'' operating point on the ROC curve would be the intersection of the ROC curve with the line in \eqref{eq:max_referral}. The $y$-intercept and the optimal operating point (and the decision threshold $\tau$), will generally change based on $T$. See~\cite{Reyna2025} for a practical application in prescreening Chagas disease, where this classification setting was explored.

\subsection{Bounded Risk Classification}
False positives and false negatives are both costly. False positives impose operational costs, require excess human (expert) interventions, and may lead to a loss of trust in AI-supported systems. False negatives may result in delayed or missed treatments, and other critical failures. While we ideally seek to minimize both, as discussed in previous sections, there is always a trade-off. In practice, this trade-off can be formulated as a risk analysis problem aiming to minimize a weighted cost of false positives and false negatives.

Assuming the risk is modeled as a linear combination of false negatives and false positives, we seek to satisfy the constraint 
$\alpha \mathrm{FP} + \beta \mathrm{FN} \leq C$,
where $C$ is the maximum allowable cost, and $\alpha$ and $\beta$ are the cost factors associated with false positives and false negatives, respectively. Using \eqref{eq:eval_metrics}, this constraint can be rewritten as $\alpha N \mathrm{fpr}(\tau) + \beta P \mathrm{fnr}(\tau) \leq C$, or, in terms of the ROC variables:
\begin{equation}
    \alpha \pi_n \mathrm{fpr}(\tau) + \beta \pi_p \bigl(1 - \mathrm{tpr}(\tau)\bigr) \leq c
\label{eq:max_risk}
\end{equation}
where $c := C / T$ is the population-size normalized cost. As shown in Fig.~\ref{fig:roc_design_plane}, in the ROC space, \eqref{eq:max_risk} defines a region above a line with positive slope $(\alpha \pi_n) / (\beta \pi_p)$ and $y$-intercept $1 - c / (\beta \pi_p)$.

\subsection{Cost-Aware Multiobjective Classification}
Another practical scenario is the case where a real-world socioeconomic cost is associated to the false positives and negatives. Missing a positive case may result in significant downstream consequences---such as a delayed diagnosis in healthcare, which could lead to disease progression, higher treatment costs, and in some cases, irreversible harm or death. These outcomes can not only impact the patient but also result in financial liabilities for the hospital, regulatory penalties, and reputational damage. On the other hand, a false positive may lead to unnecessary follow-up procedures, increased workload for clinicians, patient anxiety, and inflated operational costs. In sectors like finance or insurance, false negatives might mean undetected fraud, while false positives could flag legitimate customers, affecting user trust and potentially leading to lawsuits.

These real-world costs, though difficult to quantify precisely, highlight the need for explicitly modeling the trade-offs between model performance and annotation cost. One way to address this is to quantify and integrate ML-based design metrics and socioeconomic factors into bi-objective or multiobjective optimization frameworks. For example, in classifier design one could maximize a cost function such as
\begin{equation}
C = (1 - \lambda) \times \mathrm{performance} + \lambda \times (1 - \mathrm{cost}),
\end{equation}
where ``performance'' corresponds to any of the typical ML design metrics such as AUROC, accuracy, $F_\beta$ score, etc., $\mathrm{cost} \in [0, 1]$ is a quantification of the socioeconomic costs of false positive or negative cases, and $\lambda$ is a regularization parameter. The result of such a bi-objective optimization process is a set of solutions that form what is known as the \textit{Pareto front}~\cite{naidu2018optimal,Sameni2022}---a curve or surface representing the set of models for which no objective (e.g., performance or cost) can be improved without degrading the other. Each point on the Pareto front corresponds to a different trade-off between annotation or socioeconomic cost and predictive performance. This enables stakeholders to make principled decisions about model selection based on their operational constraints or risk tolerance. For instance, a healthcare provider might prioritize minimizing false negatives due to their clinical consequences, while an insurance company might select a model that optimally balances detection of fraudulent claims against the customer experience or potential lawsuits. Such multiobjective ML design landscapes have been explored in the applied ML literature~\cite{PerezAlday2022,Reyna2023,Sameni2022}.

In multiclass classification problems, the cost of false labels can vary across classes because some diagnoses have similar downstream actions. For example, misclassifying one benign cardiac arrhythmia as another may have little clinical consequence, whereas confusing a life-threatening condition like ventricular tachycardia with a benign rhythm could result in delayed treatment and serious harm to the patient~\cite{PerezAlday2020}. In such cases, the cost becomes a class-dependent matrix, and the optimization must account for asymmetric risks, further reinforcing the significance of cost-sensitive and multiobjective learning approaches in real-world classifier deployments.

\section{Model Calibration}
Although we have focused on discrimination metrics such as ROC and PR, many real-world applications require that a classifier's scores be interpreted as probabilities or likelihoods. For example, given a specific phenotype (feature, measurement) observed from an individual (sample point), what is the probability that the sample point belongs to the positive or negative class? The process of mapping scores to actual probabilities is known as \textit{calibration}.

For instance, if a classifier assigns a sample point $s$ a score $x(s) \approx 0.7$, then ideally, among all points assigned similar scores, approximately 70\% should belong to the positive class. This reflects proper calibration, where the output scores can be interpreted as probabilities. In practice, proper calibration is a long-term process that may be performed after a model is deployed and evaluated on large real-world data with confirmed outcomes. It can be assessed using reliability diagrams or proper scoring rules (e.g., Brier score or log loss) that compare predicted probabilities against empirical frequencies. If $x(s)$ is not inherently calibrated, post-hoc techniques such as isotonic regression or Platt scaling can adjust the raw outputs without retraining the classifier. Even a model with favorable ROC or PR characteristics may still require calibration to ensure its score outputs reflect true class probabilities in practical settings.

\section{Case Studies}
\label{sec:case_studies}
\subsection{A Random Classifier}
\label{sec:random_classifier}
A trivial corner case for \eqref{eq:roc_curve} is when $f_n(\cdot) = f_p(\cdot)$, i.e., the distribution of the score is not affected by the sample point's class label. We can think of this as a degenerate case where the classifier is unable to distinguish between the two classes. Therefore, $F_n(\cdot) = F_p(\cdot)$ (or $G(\cdot)$ is the identify function), and \eqref{eq:roc_curve} simplifies to
\begin{equation}
    \mathrm{tpr}(\tau) = \mathrm{fpr}(\tau)
\end{equation}
which corresponds to the ROC curve of a random classifier.

More generally, we can seek conditions for which the classifier performs better than chance, i.e., $\mathrm{tpr}(\tau) > \mathrm{fpr}(\tau)$, or using \eqref{eq:roc_curve}
\begin{equation}
    \mathrm{tpr}(\tau) = 1 - G \bigl(1 - \mathrm{fpr}(\tau)\bigr) > \mathrm{fpr}(\tau)
\label{eq:roc_better_than_chance}
\end{equation}
Considering that $0 \leq \mathrm{tpr}(\tau), \mathrm{fpr}(\tau) \leq 1$, and $G(\cdot)$ is monotonically increasing, $G^{-1}(\cdot)$ is also monotonically increasing and \eqref{eq:roc_better_than_chance} simplifies to $x > G(x)$, or equivalently $G^{-1}(x) > x$.

\subsection{An Ideal Classifier}
Perfect classification can be achieved when the score distributions under the negative and positive classes are entirely non-overlapping, i.e., there exists at least one decision threshold $\tau$ that places all negative sample points strictly below $\tau$ and all positive sample points strictly above it (or vice versa). Suppose $f_n(x)$ is supported only in the interval $(-\infty, \tau^*)$ and $f_p(x)$ is supported only in $[\tau^*, \infty)$. Then for any threshold $\tau$ in the open interval $\bigl(\max\,\mathrm{supp}(f_n),\; \min\,\mathrm{supp}(f_p)\bigr)$:
\begin{equation}
\mathrm{fpr}(\tau) = \int_\tau^\infty f_n(x)\,dx = 0,
\quad
\mathrm{tpr}(\tau) = \int_\tau^\infty f_p(x)\,dx = 1. \end{equation}
Hence, the ROC curve immediately jumps from the point $(\mathrm{fpr}(\tau^*)=0,\;\mathrm{tpr}(\tau)=0)$ to $(\mathrm{fpr}(\tau^*)=0,\;\mathrm{tpr}(\tau^*)=1)$, then remains at $\mathrm{tpr}(\tau^*)=1$ up to $\mathrm{fpr}(\tau^*)=1$ at the extreme right end of the ROC horizontal axis. This results in $\mathrm{AUROC}=1$. In other words, no positive sample points ever leak into the negative side for any threshold above $(-\infty,\tau^*)$, and no negative sample points appear above $\tau^*$, achieving both zero false positives and zero false negatives. Equivalently, in terms of the leakage function $G(u) = F_p\circ F_n^{-1}(u)$, since $F_n^{-1}(u)<\tau^*$ for all $u\in[0,1)$, we have $G(u)=0$ for all $u\in[0,1]$. Consequently, from \eqref{eq:auroc} we have
\begin{equation}
\int_{0}^{1} G(u)\,du = 0 \quad\Rightarrow\quad \mathrm{AUROC} = 1.    
\end{equation}
This represents the ideal yet rare case in which the classifier’s scores admit ``perfect'' separation between the negative and positive classes. In almost all practical problems, $f_n$ and $f_p$ exhibit at least some degree of overlap, so the perfect-classifier scenario does not occur. 

\subsection{Classifiers with Binormal Distribution Scores}
Consider the case where the classifier score distributions under the negative and positive classes are each Gaussian but with different means and variances~\cite[Sec 2.5]{krzanowski2009roc}:
\begin{equation}
f_n(x) = \mathcal{N}(x;\,\mu_n,\,\sigma_n^2),
\quad\quad
f_p(x) = \mathcal{N}(x;\,\mu_p,\,\sigma_p^2),
\end{equation}
with corresponding CDFs
\begin{equation}
F_n(x) = \Phi\Bigl(\frac{x - \mu_n}{\sigma_n}\Bigr),
\quad\quad
F_p(x) = \Phi\Bigl(\frac{x - \mu_p}{\sigma_p}\Bigr),
\end{equation}
where $\Phi(\cdot)$ is the standard normal CDF, with zero mean and unit standard deviation~\cite{papoulis2002probability}:
\begin{equation}
\Phi(z) = \frac{1}{\sqrt{2\pi}} \int_{-\infty}^{\,z} \exp\!\Bigl(-\tfrac{t^2}{2}\Bigr)\,dt.
\end{equation}

\subsubsection{Parametric Form of the ROC Curve}
From \eqref{eq:fpr}--\eqref{eq:tpr}, the ROC curve is given by $\bigl(\mathrm{fpr}(\tau), \mathrm{tpr}(\tau)\bigr)$ as $\tau$ varies:
\begin{equation}
\begin{aligned}
\mathrm{fpr}(\tau)
&= 1 - F_n(\tau)
= 1 - \Phi\!\Bigl(\frac{\tau - \mu_n}{\sigma_n}\Bigr),\\[3ex]
\mathrm{tpr}(\tau)
&= 1 - F_p(\tau)
= 1 - \Phi\!\Bigl(\frac{\tau - \mu_p}{\sigma_p}\Bigr).
\end{aligned}
\end{equation}
Eliminating $\tau$ yields a convenient parametric form of the ROC curve. Letting $u = (\tau - \mu_n)/\sigma_n$ we obtain
\begin{equation}
\mathrm{fpr}(u) = 1 - \Phi(u), 
\quad
\mathrm{tpr}(u) = 1 - \Phi\!\Bigl(\alpha u - b\Bigr),
\end{equation}
where
\begin{equation}
\alpha := \sigma_n/\sigma_p, \quad b:= (\mu_p - \mu_n)/\sigma_p. 
\end{equation}
Hence the ROC curve can be parameterized as
\begin{equation}
\mathrm{ROC}(u)\colon
\bigl(\mathrm{fpr}(u),\,\mathrm{tpr}(u)\bigr)
=
\bigl(1 - \Phi(u),\; 1 - \Phi(\alpha u - b)\bigr).
\end{equation}
for $u \in (-\infty,\infty)$.

\subsubsection{Analytical Expression for the Leakage Function \texorpdfstring{\boldmath$G(\cdot)$}{}}
The inverse CDF of $F_n(\cdot)$ is:
\begin{equation}
F_n^{-1}(u)
= \mu_n + \sigma_n\,\Phi^{-1}(u).
\end{equation}
Therefore,
\begin{equation}
\begin{array}{rl}
G(u)
& = F_p\bigl(F_n^{-1}(u)\bigr)
= \Phi\Bigl(\displaystyle\frac{F_n^{-1}(u) - \mu_p}{\sigma_p}\Bigr)\\[3ex]
& = \Phi\Bigl(\displaystyle\frac{\mu_n + \sigma_n\,\Phi^{-1}(u) - \mu_p}{\sigma_p}\Bigr)
=
\Phi\bigl(\alpha\Phi^{-1}(u) - b\bigr),
\end{array}
\label{eq:gaussian_G_function}
\end{equation}
Thus for two Gaussian score distributions, the function $G(u)$ is simply a composition of standard normal CDFs and their inverses. Its shape is fully determined by the ratio between the standard deviations ($\alpha$) and the normalized gap between their means ($b$). It is apparent that gaussian distributions with farther mean values and smaller standard deviations result in lower inter-class leakage.

\subsubsection{Analytical AUROC Expression}
Finally, using \eqref{eq:auroc_prob}, since the difference $x_p - x_n$ is normally distributed with $x_p - x_n \sim 
\mathcal{N}\bigl(\mu_p - \mu_n,\sigma_n^2 + \sigma_p^2\bigr)$, one obtains~\cite[Sec 2.5]{krzanowski2009roc}:
\begin{equation}
\mathrm{AUROC} = \displaystyle\Phi\left(\frac{\mu_p - \mu_n}{\sqrt{\sigma_n^2 + \sigma_p^2}}\right),
\end{equation}
which according to \eqref{eq:auroc} and \eqref{eq:gaussian_G_function} is non-evidently one minus the area under $\Phi\bigl(\alpha\Phi^{-1}(u) - b\bigr)$.

This example confirms that as the separation $(\mu_p - \mu_n)$ grows and/or the spread $(\sigma_n^2 + \sigma_p^2)$ shrinks, the AUROC increases and the classifier becomes more discriminative. By contrast, when $\mu_p = \mu_n$, the AUROC degenerates to $0.5$, or a random classifier explained in Section~\ref{sec:random_classifier}.

\begin{figure*}[tb]
    \centering

    \begin{subfigure}[t]{0.32\textwidth}
        \centering
        \includegraphics[width=\columnwidth]{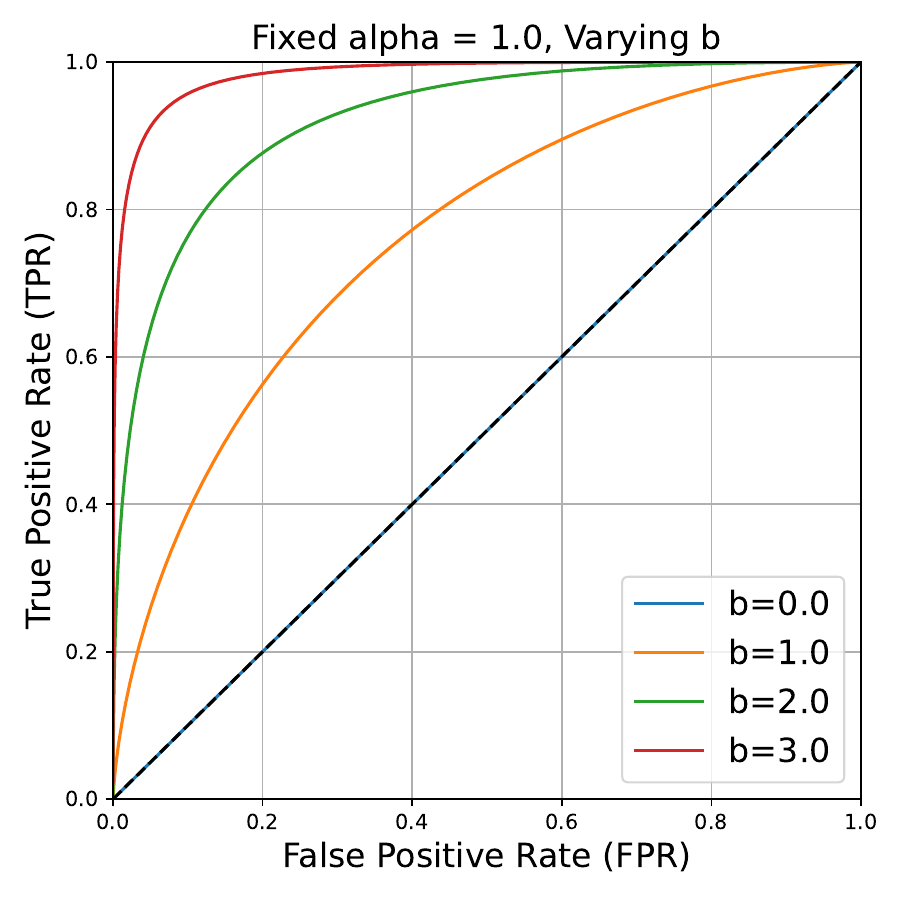}
        \caption{Fixed $\alpha = 1.0$, varying $b$}
        \label{fig:roc_fixed_alpha_varying_b}
    \end{subfigure}
    \hfill
    \begin{subfigure}[t]{0.32\textwidth}
        \centering
        \includegraphics[width=\columnwidth]{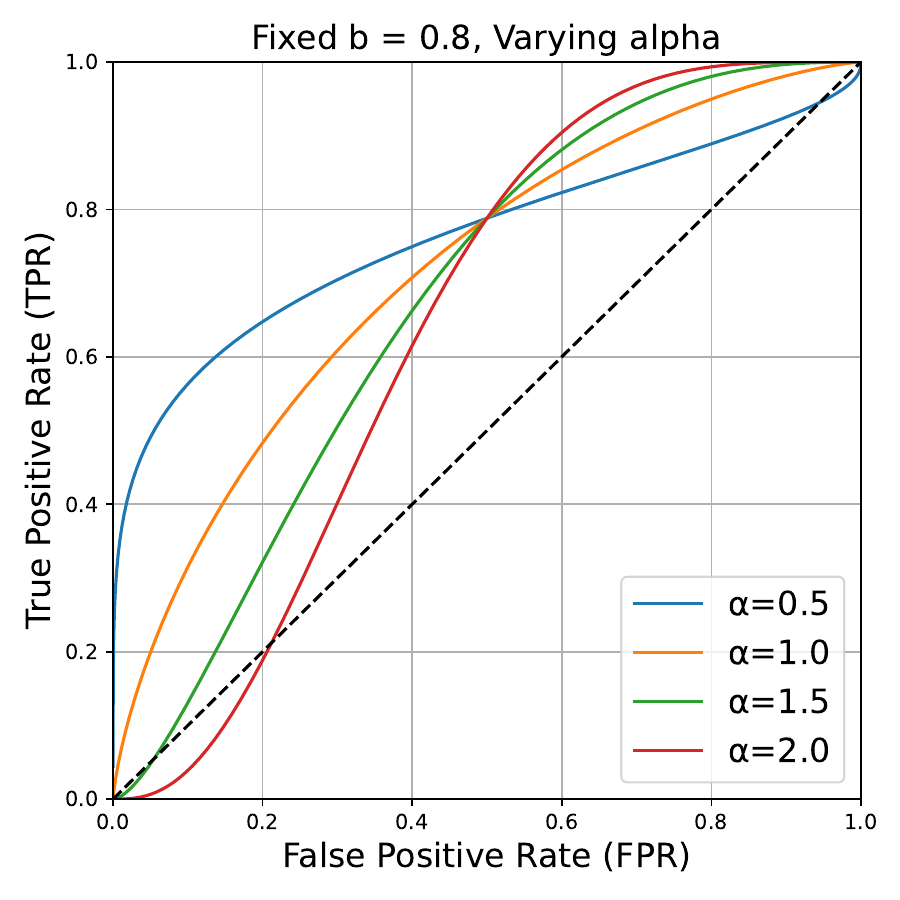}
        \caption{Fixed $b = 0.8$, varying $\alpha$}
        \label{fig:roc_fixed_b_varying_alpha}
    \end{subfigure}
    \hfill
    \begin{subfigure}[t]{0.32\textwidth}
        \centering
        \includegraphics[width=\columnwidth]{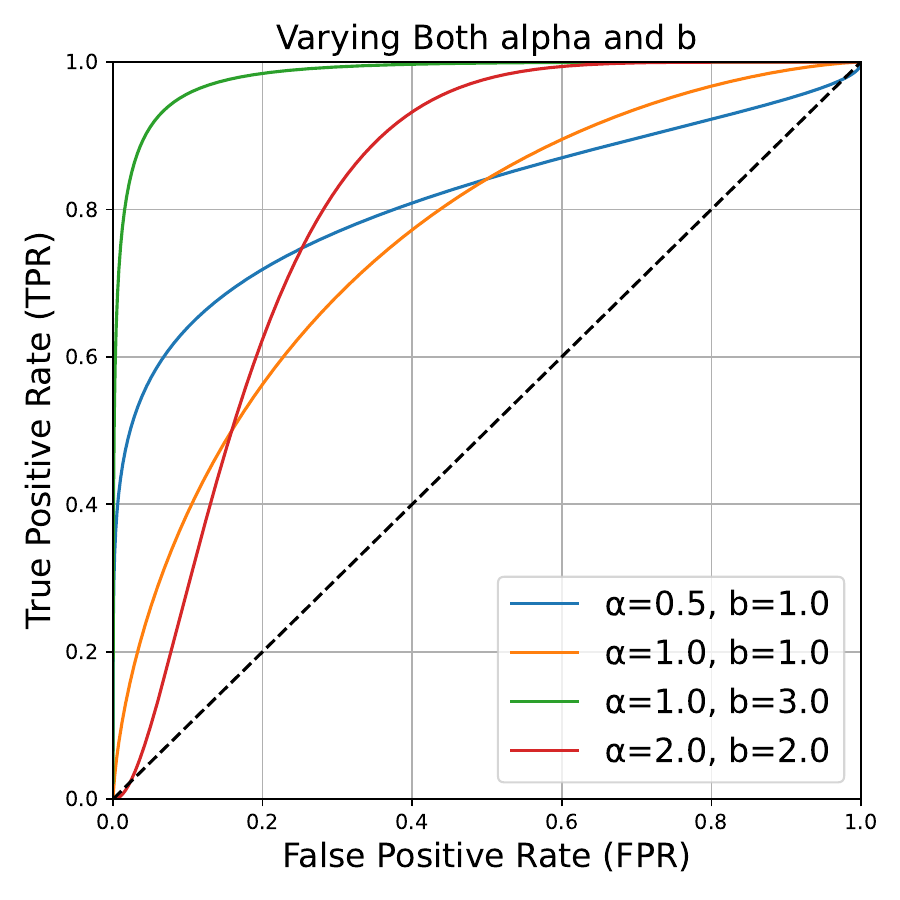}
        \caption{Varying both $\alpha$ and $b$}
        \label{fig:roc_varying_alpha_and_b}
    \end{subfigure}

    \caption{ROC curves for Gaussian score classifiers under different parameter settings. (a) Varying $b$ with fixed $\alpha$; (b) Varying $\alpha$ with fixed $b$; (c) Joint variation of both parameters.}
    \label{fig:roc_case_study}
\end{figure*}

\paragraph{Numeric Examples} To study the impact of score distribution parameters on ROC geometry, we conduct three sets of experiments using Gaussian-distributed scores. In the first experiment (Fig.~\ref{fig:roc_fixed_alpha_varying_b}), the variance ratio is fixed at $\alpha = 1.0$ while the normalized mean separation $b$ is varied across $\{0.0, 1.0, 2.0, 3.0\}$. This setup isolates the effect of increasing class separation on ROC performance. In the second experiment (Fig.~\ref{fig:roc_fixed_b_varying_alpha}), the mean separation is fixed at $b = 0.8$ and $\alpha$ is varied across $\{0.5, 1.0, 1.5, 2.0\}$ to examine the influence of score dispersion between classes. In the third case (Fig.~\ref{fig:roc_varying_alpha_and_b}), both $\alpha$ and $b$ are varied to study their combined effect on classifier discriminability. All plots exhibit the expected trends: increasing $b$ enhances ROC performance, while deviations in $\alpha$ modulate the steepness and shape of the curves, reflecting changes in overlap between the class-conditional score distributions.

\begin{figure}[tb]
    \centering
    \includegraphics[width=0.32\textwidth]{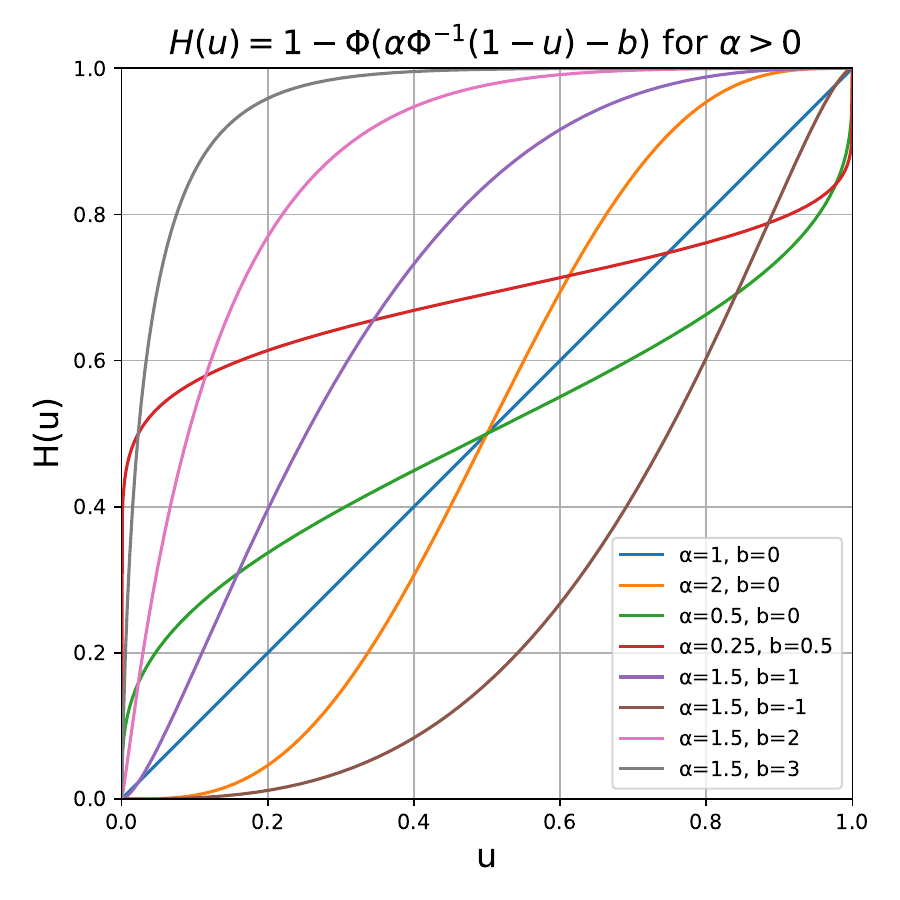}
    \caption{
        Generic ROC family generated from the transformation $H(u) = 1 - \Phi(\alpha \Phi^{-1}(1 - u) - b)$ using different values of $\alpha=\sigma_n/\sigma_p > 0$ and $b=(\mu_p - \mu_n)/\sigma_p$. When $\alpha = 1$ and $b = 0$, the function reduces to the identity line (random classifier). As $\alpha$ increases, the curves become steeper, representing classifiers with more confident separation between positive-negative classes. Increasing $b$ helps improve separability of the two classes. Negative $b$ implies that the class labels or decision rule should be switched (positive class being assigned to lower thresholds).
    }
    \label{fig:generic_roc_case_study}
\end{figure}

Fig.~\ref{fig:generic_roc_case_study} shows generic ROC curves corresponding to two-Gaussian distributed scores, generated using the expression $H(u) = 1 - \Phi(\alpha \Phi^{-1}(1 - u) - b)$, which results from combining \eqref{eq:roc_curve} and \eqref{eq:gaussian_G_function}. Each curve corresponds to a different combination of $(\alpha, b)$, demonstrating the impact of the Gaussian distribution parameters on the ROC shape. When $\alpha = 1$ and $b = 0$, the function reduces to the identity line, corresponding to a random classifier. As $\alpha$ increases, the curves become steeper, representing classifiers with more confident separation between positive-negative classes. Shifting $b$ to positive or negative values translates the curve right or left, respectively, modifying the threshold at which positive predictions occur.

\begin{figure*}[tb]
    \centering
    \begin{subfigure}[t]{0.32\textwidth}
        \centering
        \includegraphics[width=\textwidth]{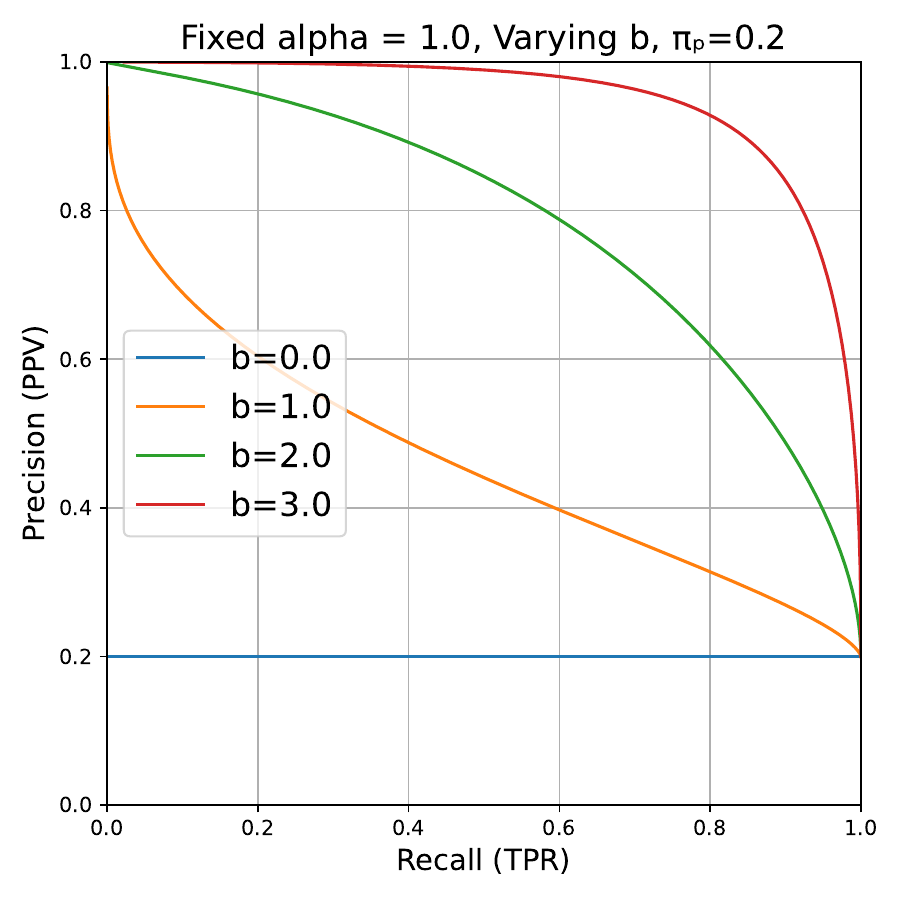}
        \caption{$\pi_p = 0.2$, $\alpha = 1.0$, varying $b$}
        \label{fig:prc_fixed_alpha_varying_b}
    \end{subfigure}
    \hfill
    \begin{subfigure}[t]{0.32\textwidth}
        \centering
        \includegraphics[width=\textwidth]{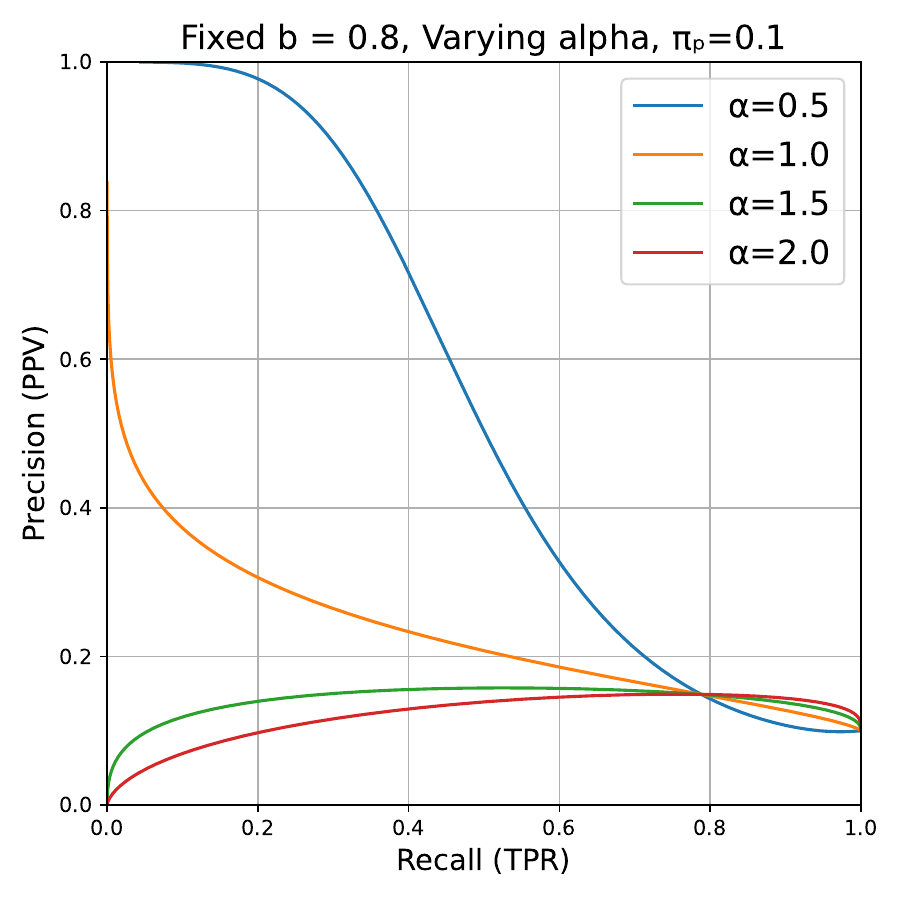}
        \caption{$\pi_p = 0.1$, $b = 0.8$, varying $\alpha$}
        \label{fig:prc_fixed_b_varying_alpha}
    \end{subfigure}
    \hfill
    \begin{subfigure}[t]{0.32\textwidth}
        \centering
        \includegraphics[width=\textwidth]{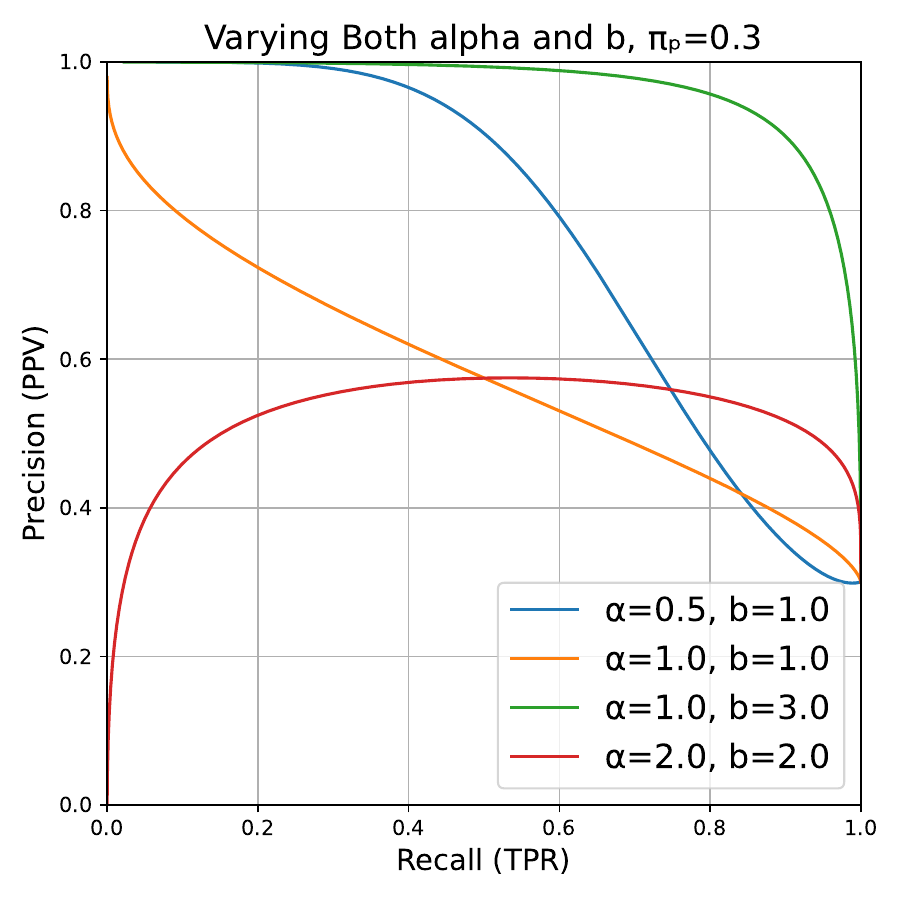}
        \caption{$\pi_p = 0.3$, varying both $\alpha$ and $b$}
        \label{fig:prc_varying_alpha_and_b}
    \end{subfigure}
    \caption{PR curves for binomial positive and negative class scores under varying classifier parameters and class prevalence ratios. Lower $\pi_p/\pi_n$ ratios result in reduced precision, despite similar recall performance.}
    \label{fig:prc_case_study}
\end{figure*}

To complement the ROC analysis, we conduct three sets of experiments to explore how class prevalence influences the shape and behavior of PR curves. In all cases, the score distributions are modeled as Gaussians with varying mean and variance parameters, described above. Unlike ROC curves, PR curves are sensitive to class imbalance. Therefore, our experiments vary the positive-to-negative class prevalence ratios at $\pi_p = \{0.2, 0.1, 0.3\}$ to study this dependency. In the first scenario (Fig.~\ref{fig:prc_fixed_alpha_varying_b}), the variance ratio is fixed at $\alpha = 1.0$, while the mean separation $b$ varies over $\{0.0, 1.0, 2.0, 3.0\}$. The second experiment (Fig.~\ref{fig:prc_fixed_b_varying_alpha}) fixes $b = 0.8$ and varies $\alpha \in \{0.5, 1.0, 1.5, 2.0\}$, demonstrating how dispersion affects PR behavior. The third experiment (Fig.~\ref{fig:prc_varying_alpha_and_b}) tests four $(\alpha, b)$ combinations under different class ratios. As expected, increasing class imbalance (i.e., decreasing $\pi_p/\pi_n$) shifts precision downward, even when recall remains high.

\begin{figure*}[tb]
    \centering
    \begin{subfigure}[t]{0.24\textwidth}
        \centering
        \includegraphics[width=\textwidth]{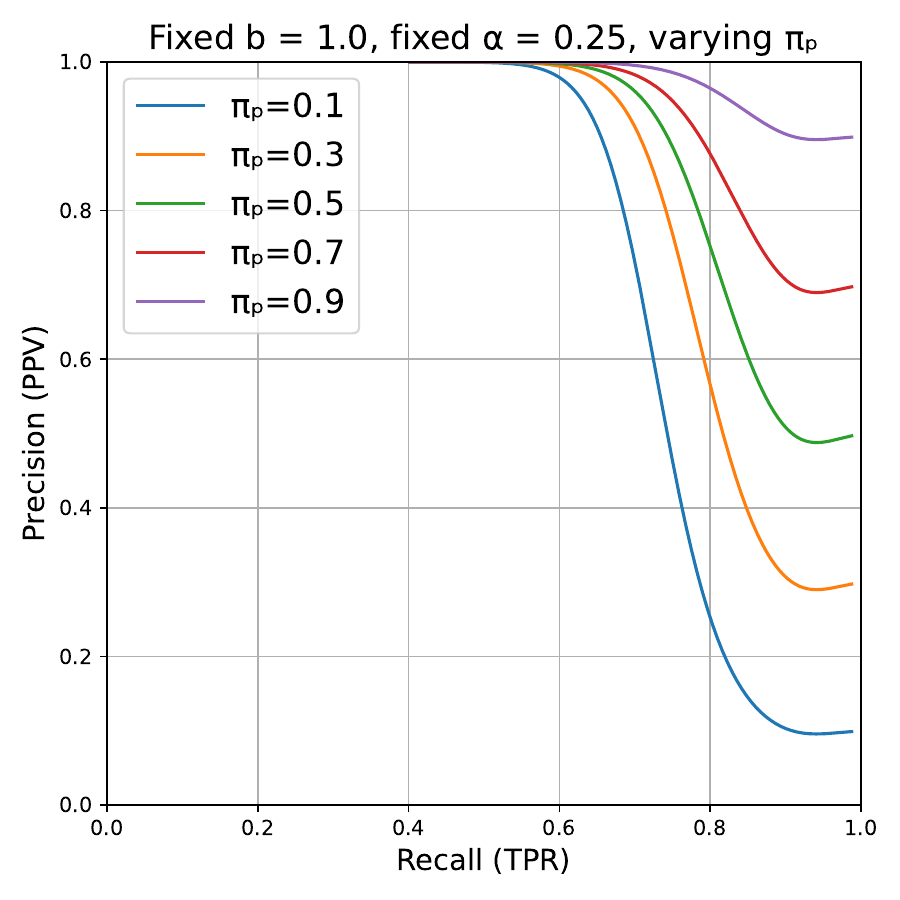}
        \caption{$b = 1$, $\alpha = 0.25$, vary $\pi_p$}
        \label{fig:prc_fixed_b_and_alpha_varying_prevalence_s1}
    \end{subfigure}
    \hfill
    \begin{subfigure}[t]{0.24\textwidth}
        \centering
        \includegraphics[width=\textwidth]{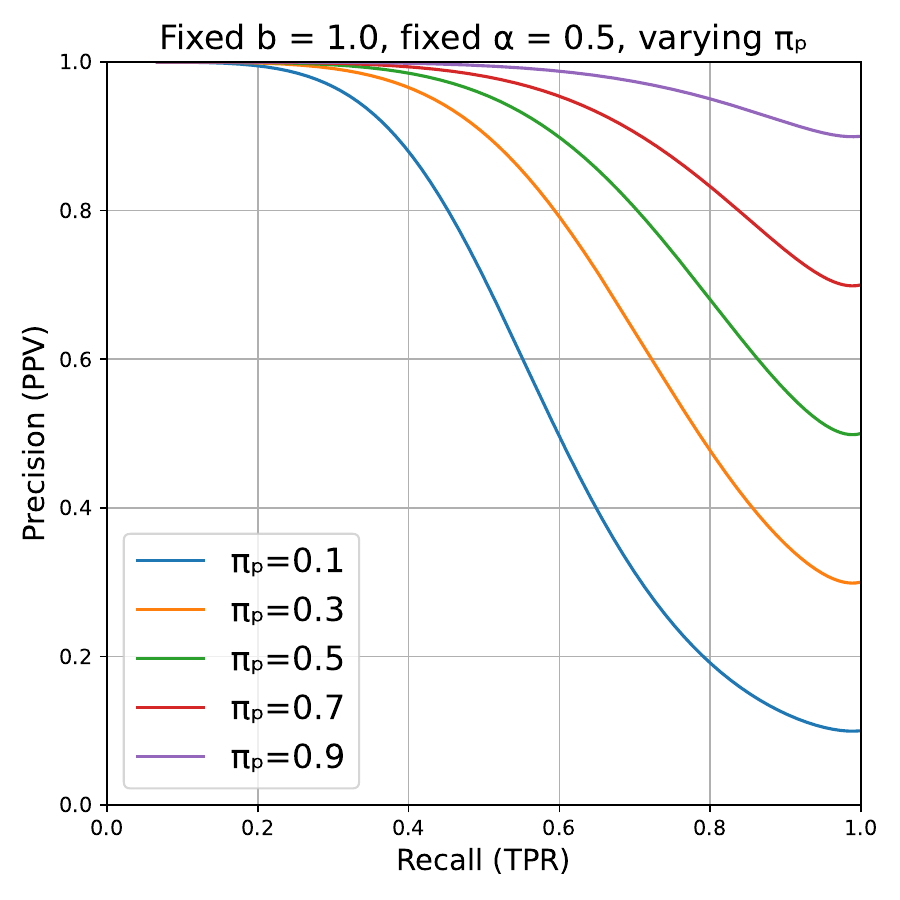}
        \caption{$b = 1$, $\alpha = 0.5$, vary $\pi_p$}
        \label{fig:prc_fixed_b_and_alpha_varying_prevalence_s2}
    \end{subfigure}
    \hfill
    \begin{subfigure}[t]{0.24\textwidth}
        \centering
        \includegraphics[width=\textwidth]{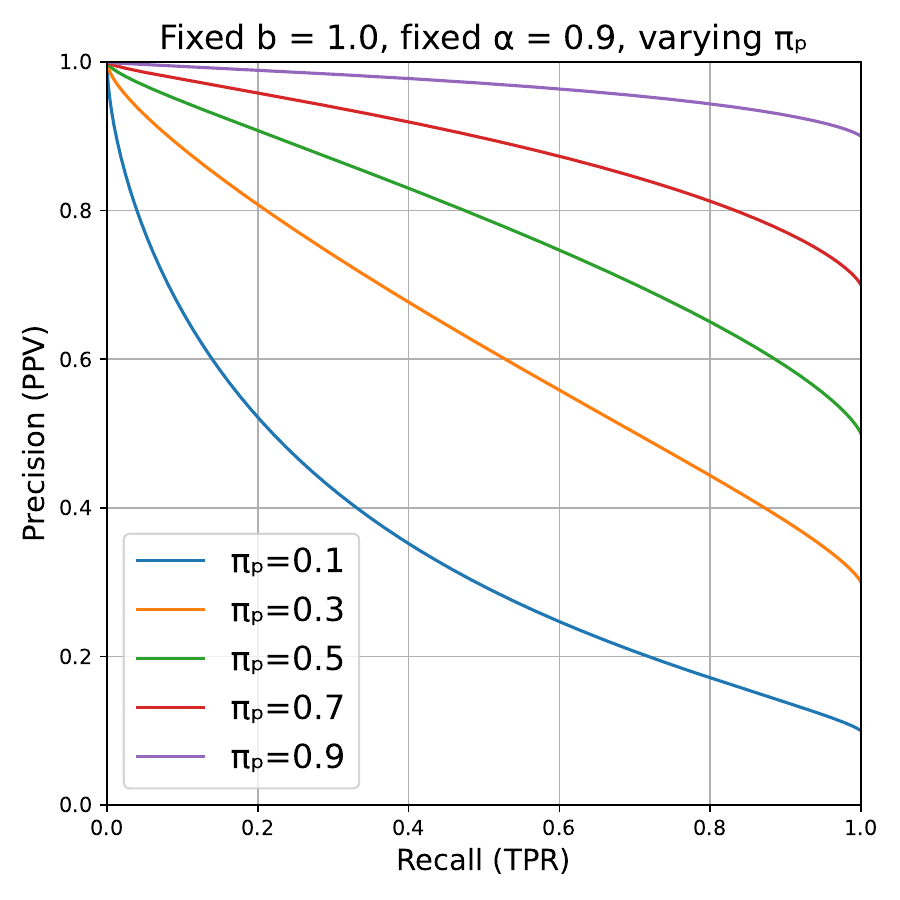}
        \caption{$b = 1$, $\alpha = 0.9$, vary $\pi_p$}
        \label{fig:prc_fixed_b_and_alpha_varying_prevalence_s3}
    \end{subfigure}
    \hfill
    \begin{subfigure}[t]{0.24\textwidth}
        \centering
        \includegraphics[width=\textwidth]{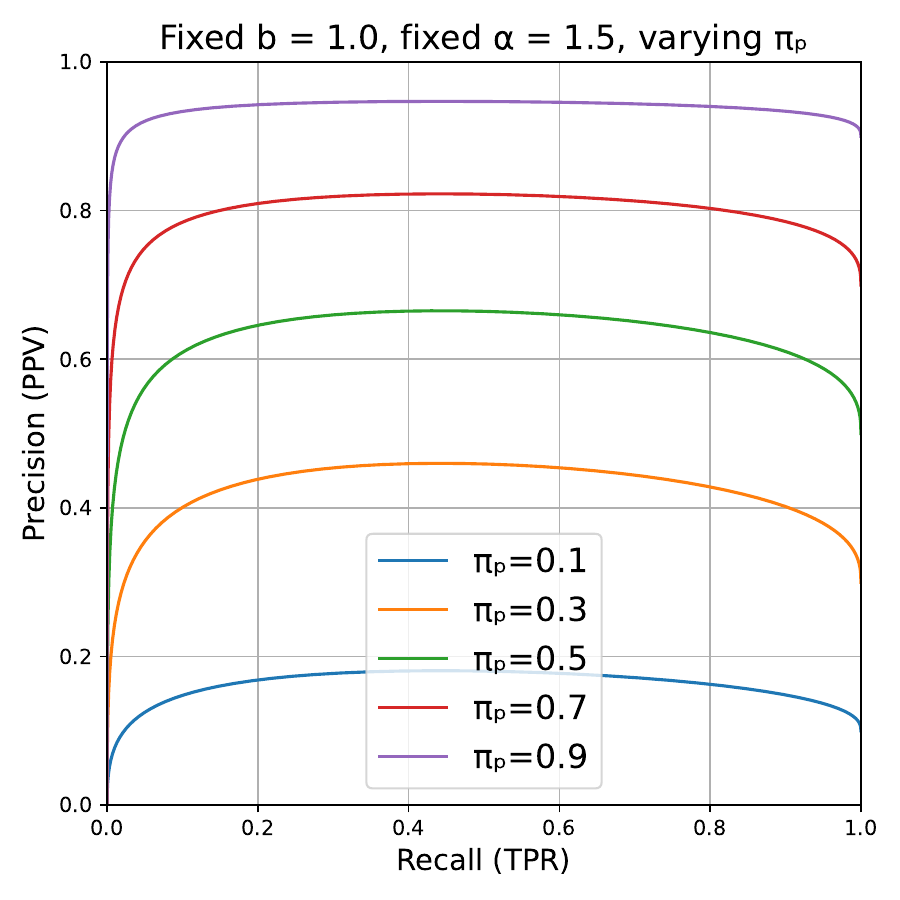}
        \caption{$b = 1$, $\alpha = 1.5$, vary $\pi_p$}
        \label{fig:prc_fixed_b_and_alpha_varying_prevalence_s4}
    \end{subfigure}
    \caption{PR curves for binomial positive and negative class scores under fixed values of $b$ and $\alpha$ and varying positive class prevalence ratios.}
    \label{fig:prc_case_study_varying_pi}
\end{figure*}

Fig.~\ref{fig:prc_case_study_varying_pi} shows other examples where $b$ and $\alpha$ are fixed as $\pi_p$ varies between $\{0.1, 0.3, 0.5, 0.7, 0.9\}$. We can see the diversity of PR curves.

%////////////////////////////////////
\section*{Acknowledgment}
The author sincerely thanks \href{https://orcid.org/0000-0002-9172-8413}{Dr. Amin Zollanvari} from the Department of Electrical and Computer Engineering at Nazarbayev University, and \href{https://orcid.org/0000-0003-4688-7965}{Dr. Matthew A. Reyna} from the Department of Biomedical Informatics at Emory University, for their valuable feedback, fruitful discussions and insightful comments on this work.

%////////////////////////////////////
\appendices
% \section*{Appendix}
\section{Proof of Probabillistic Interpretation for AUROC}
\label{sec:auroc_proof}
We show that $\mathrm{AUROC} = \Pr(x_p > x_n)$, i.e., the AUROC is equal to the probability that a randomly drawn positive sample point receives a higher score than a randomly drawn negative one (cf.~\cite[Sec.~2.4]{krzanowski2009roc}). First,
\begin{equation}
\begin{array}{l}
\mathrm{AUROC}=\displaystyle\int_{-\infty}^{\infty} \mathrm{tpr}(t)\, d\bigl[\mathrm{fpr}(t)\bigr]=\displaystyle\int_0^1 \mathrm{tpr}(\mathrm{fpr}^{-1}(s))\, ds \\[3ex]
= \displaystyle\int_{-\infty}^{\infty} \mathrm{tpr}(t)\, f_n(t)\, dt= \displaystyle\int_{-\infty}^{\infty} \left[ \int_t^\infty f_p(x)\, dx \right] f_n(t)\, dt \\[3ex]
= \displaystyle\int_{-\infty}^{\infty} f_p(x)\, \left[ \int_{-\infty}^x f_n(t)\, dt \right] dx = \displaystyle\int_{-\infty}^{\infty} f_p(x)\, F_n(x)\, dx \\[3ex]
= \Pr(x_p > x_n).
\end{array}
\end{equation}
% \begin{equation}
% \mathrm{AUROC} = \Pr(x_p > x_n) = \int_{-\infty}^{\infty} f_p(x)\, F_n(x)\, dx
% \end{equation}
where we differentiated \eqref{eq:fpr} and a change of variable, to go from the second to the third line. Next, assuming $x_p \sim f_p(x)$ and $x_n \sim f_n(x)$ are independent, we obtain:
\begin{equation}
\begin{array}{rl}
    \Pr(x_p > x_n) & =  \displaystyle\iint_{x_p > x_n} f_p(x_p) f_n(x_n)\, dx_p\, dx_n\\[3ex]
    =  & \displaystyle\int_{-\infty}^{\infty} f_p(x_p) \left[ \int_{-\infty}^{x_p} f_n(x_n)\, dx_n \right] dx_p \\[3ex]
    & = \displaystyle\int_{-\infty}^{\infty} f_p(x) F_n(x)\, dx.
\end{array}
\end{equation}

\section{Kullback–Leibler Divergence vs Positive-to-Negative Class Leakage Function}
\label{sec:kl_leakage_derivation}
We derive the Kullback–Leibler (KL) divergence between the class-conditional score distributions $f_p(x)$ and $f_n(x)$ using the leakage function $G(u) = F_p(F_n^{-1}(u))$ introduced in~\eqref{eq:G}.

Starting from the standard KL divergence definition:
\begin{equation}
D_{\mathrm{KL}}(f_p \| f_n) := \int_{-\infty}^{\infty} f_p(x) \log \left( \frac{f_p(x)}{f_n(x)} \right) dx,
\end{equation}
we apply the change of variable $x = F_n^{-1}(u)$, where $u = F_n(x)$, so that
\begin{equation*}
dx = \frac{1}{f_n(F_n^{-1}(u))}\, du.    
\end{equation*}
Substituting into the integral gives:
\begin{equation}
D_{\mathrm{KL}}(f_p \| f_n) = \int_0^1 \frac{f_p(F_n^{-1}(u))}{f_n(F_n^{-1}(u))} \log \left( \frac{f_p(F_n^{-1}(u))}{f_n(F_n^{-1}(u))} \right) du.
\end{equation}
From \eqref{eq:G}, the derivative of the leakage function is
\begin{equation}
\dot{G}(u) = \frac{f_p(F_n^{-1}(u))}{f_n(F_n^{-1}(u))}.
\label{eq:G_derivative}
\end{equation}
Substituting into the KL expression and combining with \eqref{eq:G_density} yields:
\begin{equation}
D_{\mathrm{KL}}(f_p \| f_n) = \!\! \int_0^1 \dot{G}(u)\, \log \dot{G}(u)\, du = \!\!\int_0^1 g(u)\, \log g(u)\, du.
\label{eq:kl_divergence_diff_entropy}
\end{equation}
which is the negative \textit{differential entropy} of $g(\cdot)$\footnote{Remember that $G(\cdot)$ and $g(\cdot)$ are only defined over $[0, 1]$.}. This expression relates the information-theoretic divergence between $f_p$ and $f_n$ to the geometric derivative of the leakage function $G$, which also underlies ROC and PR curve behavior.

\paragraph{Interpretation}
The differential entropy of $g(u) = \dot{G}(u)$ provides an interpretable measure of the classifier behavior. The function $g(u)$ captures the rate at which scores of positive sample points leak into the score region corresponding to the $u$-quantile of the negative class. The differential entropy on the right hand side of \eqref{eq:kl_divergence_diff_entropy} quantifies how this leakage is distributed across the decision threshold space. A high entropy value indicates that the classifier's confusion is spread over a broad range of thresholds, implying misclassifications occur across many regions of the score space. In contrast, a low entropy implies that the leakage is concentrated in a narrow range, suggesting that the classifier generally separates the classes well and only struggles in a localized region. In this way, the entropy of $g(u)$ shows whether the classifier's errors are widespread and systematic (being generally a poor classifier) or localized and potentially correctable. Since the KL divergence between the positive and negative score distributions is directly related to class separability: lower entropy corresponds to more focused leakage and higher KL divergence, indicating a more discriminative classifier.

\bibliographystyle{ieeetr}
\bibliography{references}
\end{document}